\documentclass{article}

\usepackage{longcat_style}
\usepackage{adjustbox}
\usepackage[utf8]{inputenc} %
\usepackage[T1]{fontenc}    %
\usepackage{newunicodechar}
\usepackage{hyperref}       %
\usepackage{xcolor}
\usepackage[normalem]{ulem} %
\hypersetup{
    colorlinks=true,      %
    linkcolor=blue,      %
    urlcolor=blue,       %
    citecolor=blue,      %
    linkbordercolor=blue, %
    urlbordercolor=blue,
    citebordercolor=blue,
    pdfborderstyle={/S/U/W 1}, %
}

\usepackage{float}
\usepackage{placeins}
\usepackage{url}            %
\usepackage{booktabs}       %
\usepackage{amsfonts}       %
\usepackage{nicefrac}       %
\usepackage{microtype}      %
\usepackage{lipsum}		%
\usepackage{graphicx}
\usepackage{natbib}
\usepackage{doi}
\usepackage{amsmath}
\usepackage{amssymb} %
\usepackage{xspace}
\usepackage{enumitem}
\usepackage{multirow}
\usepackage{subcaption} 
\usepackage{makecell}
\usepackage{hyperref}
\usepackage[capitalise]{cleveref}
\usepackage{pifont}
\usepackage[inkscapelatex=false]{svg}
\usepackage{subcaption}
\usepackage[most]{tcolorbox} %
\usepackage{pgfplots}
\pgfplotsset{compat=1.18}
\usepgfplotslibrary{groupplots}

\newtcolorbox{summarybox}{
    colback=blue!5!white,    %
    colframe=black!75,       %
    arc=6pt,                 %
    boxrule=1pt,             %
    left=8pt,                %
    right=8pt,               %
    top=8pt,                 %
    bottom=8pt,              %
    enhanced,                %
}

\setlist[itemize]{leftmargin=*}
\setlist[enumerate]{leftmargin=*}
\setlist[description]{leftmargin=*}

\newcommand{\longcat}{LongCat-Flash-Lite\xspace}

\newcommand{\kmax}{K}

\newcommand{\mean}{\text{mean}}

\definecolor{midnightgreen}{rgb}{0.0, 0.29, 0.33}

\title{LongCat Sparse Attention: Taming the Lightning via Streaming-aware Hierarchical Cross-Layer Indexing}

\author{ Wen Zan, Jiaqi Zhang\footnotemark[1], Jianchao Tan, Hong Liu, Cunguang Wang,\\
\textbf{Xiang Li, Duyue Ma, Guanyu Wu, Yifan Lu, Fengcun Li,} \\
\textbf{ Yerui Sun, Peng Pei, Yuchen Xie, Xunliang Cai} \\
\\
\textbf{Meituan LongCat Team}
}

\renewcommand{\headeright}{\raisebox{-0.2\height}{\includegraphics[height=1.8em]{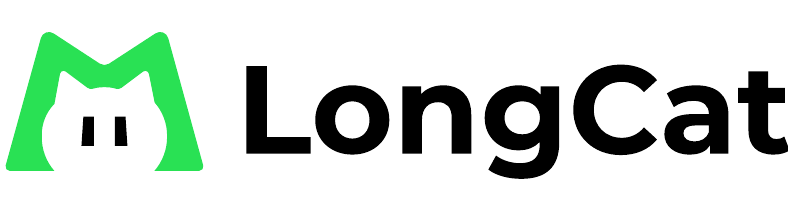}}}
\renewcommand{\shorttitle}{LongCat Sparse Attention}

\begin{document}
\maketitle

\footnotetext[1]{Corresponding authors: zhangjiaqi39@meituan.com}

\begin{abstract}
DeepSeek Sparse Attention (DSA) enables efficient long-context modeling through its Lightning Indexer. However, practical deployment remains constrained by the indexer’s expensive $\mathcal{O}(L^2)$ scoring overhead and the hardware-inefficient, discontinuous memory-access patterns induced by its outputs. To address these system-level bottlenecks, we introduce \textbf{LongCat Sparse Attention} (LSA), a hardware–algorithm co-designed framework comprising three complementary and orthogonal strategies: (1) {Streaming-Aware Indexing}, which selectively converts scattered KV entries into hardware-aligned contiguous layouts to enable coalesced HBM access; (2) {Cross-Layer Indexing}, which amortizes indexing overhead by reusing the results produced by a single layer across consecutive layers, supported by cross-layer distillation; and (3) {Hierarchical Indexing}, which adopts a coarse-to-fine scoring scheme to progressively narrow the candidate set for each query, thereby substantially reducing indexing computation. Extensive scaling experiments, ranging from 69B-A3B to 560B-A27B models, demonstrate that LSA consistently achieves performance on par with full attention across both general-purpose and long-context benchmarks. Moreover, LSA supports native training with context lengths of up to one million tokens and underpins the development of \textbf{LongCat-2.0} (1.6T-A48B). To facilitate further research, we also introduce and open-source \textbf{LongCat-Flash-Lite-Sparse} (69B-A3B), which integrates LSA into LongCat-Flash-Lite and incorporates an updated long-context training corpus.
\end{abstract}

\begin{figure}[h]
    \centering
    \includegraphics[width=1.0\linewidth]{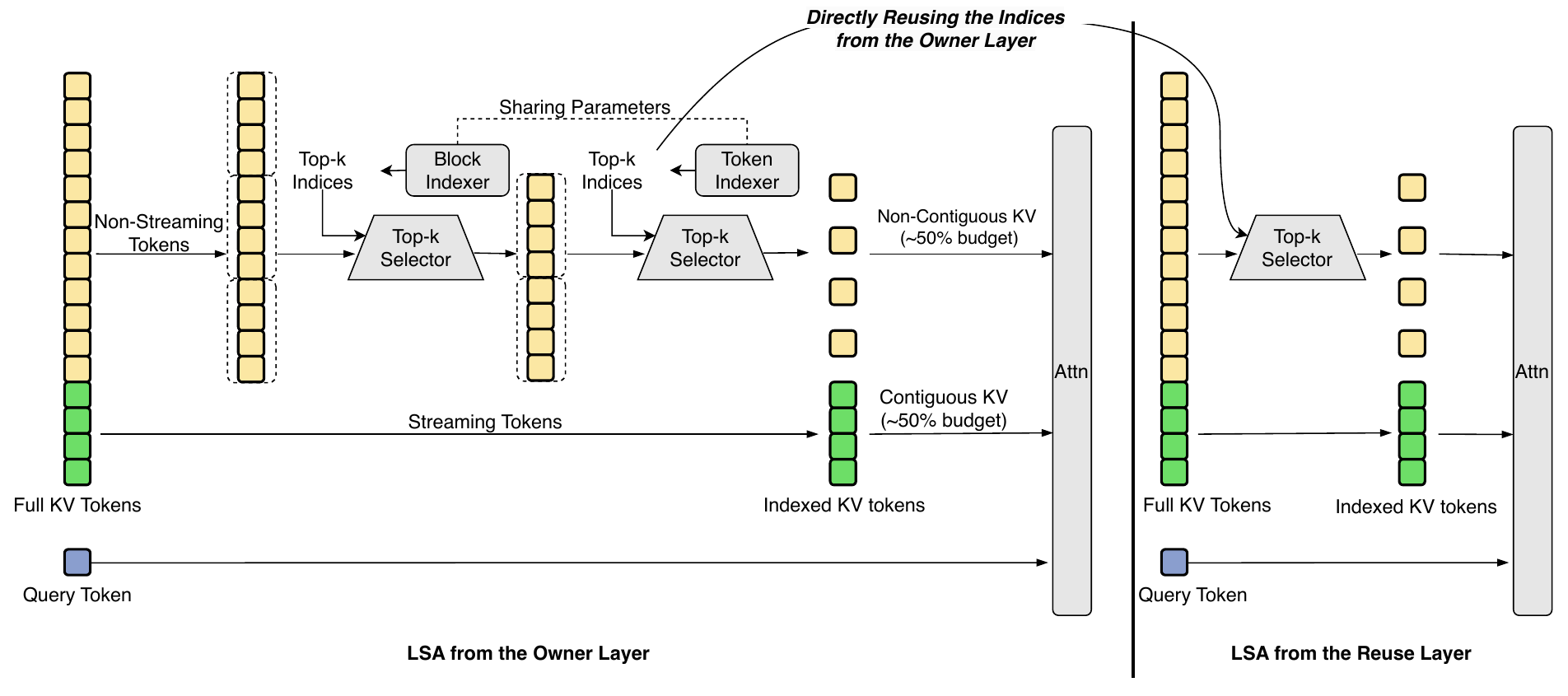}
    \caption{The architecture of the proposed LongCat Sparse Attention (LSA), featuring the streaming-aware hierarchical cross-layer indexer. (Sink tokens are omitted for clarity.)}
    \label{fig:overview}
\end{figure}

\clearpage
\tableofcontents
\clearpage

\section{Introduction}
\label{sec:intro}
Long-context capabilities are increasingly essential for modern large language models (LLMs), underpinning applications such as repository-level agentic coding, long-horizon task execution, and extended reasoning. However, scaling LLMs to long contexts remains fundamentally constrained by the quadratic $\mathcal{O}(L^2)$ complexity of standard self-attention~\citep{vaswani2017attention}, where $L$ denotes the sequence length. Sparse attention mitigates this bottleneck by restricting each query to a subset of key-value (KV) tokens, but its effectiveness depends critically on selecting the most relevant tokens both accurately and efficiently. DeepSeek Sparse Attention (DSA)~\citep{deepseek2025dsa} addresses this challenge with a dedicated \emph{Lightning Indexer}, which scores prefix tokens for each query and performs fine-grained, token-level selection. DSA thereby achieves near-lossless quality relative to full attention and has been adopted by production-scale foundation models, including DeepSeek-V3.2~\citep{deepseek2025v32} and GLM-5~\citep{zeng2026glm5}.

Despite its advantages over full attention at long sequence lengths, DSA still leaves substantial room for efficiency improvement. Each DSA layer sequentially executes two operators: the \emph{Lightning Indexer}, which scores prefix tokens and selects a budget of $\kmax$ candidates, and the \emph{core sparse attention} operator, which attends only to the selected subset. Our profiling reveals two corresponding system-level bottlenecks: \emph{Indexer Output Discontiguity} and \emph{Indexer High Overhead}.

\emph{Indexer Output Discontiguity} arises because dynamically selected indices require the core sparse-attention operator to gather KV vectors through scattered, non-coalesced memory accesses. Consequently, despite its modest $\mathcal{O}(L\kmax)$ computational complexity, the operator becomes severely memory-bound and achieves only ${\sim}4.5\%$ of peak HBM bandwidth on our AI accelerators. \emph{Indexer High Overhead}, meanwhile, stems from scoring the entire prefix for every query, which retains a quadratic $\mathcal{O}(L^2)$ complexity. Although DSA reduces the indexer's constant factors through fewer heads, smaller head dimensions, and FP8 quantization, its quadratic cost increasingly dominates layer latency as the context grows, accounting for up to $90\%$ of the per-layer latency at a 1024K context length. We characterize both bottlenecks in detail in \cref{sec:profiling}.

To address \emph{Indexer Output Discontiguity}, we introduce \emph{Streaming-Aware Indexing (SI)}. SI allocates part of the sparse attention budget to predictable and contiguous regions, specifically a fixed attention sink and a local sliding window, while using the remaining budget for dynamic token-level selection. This design is motivated by the streaming patterns observed in full attention~\citep{xiao2023streamingllm,xiao2024duoattention}. By combining dynamically selected tokens with contiguous local regions, SI improves memory coalescing and restores high HBM throughput without compromising long-context accuracy.

To reduce \emph{Indexer High Overhead}, we introduce two complementary mechanisms: \emph{Cross-Layer Indexing (CLI)} and \emph{Hierarchical Indexing (HI)}. CLI amortizes indexing computation across model depth. Motivated by the observation that salient tokens are often highly consistent across adjacent layers~\citep{yang2024tidaldecode,deshmukh2025kascade,gao2026hysparse}, CLI partitions consecutive layers into groups of size $N$. The first layer in each group, termed the \emph{owner layer}, executes the indexer, while the remaining \emph{reuse layers} share its selected token set. Naively reusing the owner layer's indices, however, leads to substantial quality degradation. We therefore introduce a cross-layer distillation objective that trains the owner indexer to capture tokens salient to the entire group. This reduces indexing computation to approximately $1/N$ of that required by standard DSA while preserving model quality.

HI reduces the cost of each indexing pass through a coarse-to-fine selection scheme. It first uses block-level representations to retrieve the top-$M$ candidate pages and then performs fine-grained token scoring only within those pages. This reduces the per-query selection complexity from $\mathcal{O}(L)$ to $\mathcal{O}(L/P + MP)$, where $P$ denotes the page size. Unlike SI and CLI, which are incorporated during both training and inference, HI is a training-free, plug-and-play module that can be applied directly at inference time.

Together, SI, CLI, and HI form \textbf{LongCat Sparse Attention (LSA)}, a co-designed sparse-attention framework whose components target complementary bottlenecks and can be composed seamlessly (\cref{fig:overview}). We evaluate LSA at two scales within the LongCat family: LongCat-Flash-Lite (69B-A3B) and LongCat-Flash (560B-A27B). Across both scales, LSA consistently matches full attention on general, reasoning, coding, and long-context benchmarks. Moreover, its substantially improved training efficiency over full MLA enables the training of \textbf{LongCat-2.0} (1.6T-A48B)\footnote{\url{https://huggingface.co/meituan-longcat/LongCat-2.0}} with context lengths of up to one million tokens under a limited compute budget.

Building on LSA, we further extend the long-context training of LongCat-Flash-Lite~\citep{liu2026scaling}, scaling its native context length to one million tokens. The resulting model, \textbf{LongCat-Flash-Lite-Sparse}, achieves substantially higher long-context inference efficiency than its dense predecessor while exhibiting stronger agentic capabilities.

Our main contributions are summarized as follows:
\begin{itemize}
    \item We systematically profile DSA and identify two major efficiency bottlenecks: \emph{Indexer Output Discontiguity} and \emph{Indexer High Overhead} (\cref{sec:profiling}).

    \item We propose LSA, which addresses these bottlenecks through three complementary mechanisms: Streaming-Aware Indexing for hardware-friendly memory access, Cross-Layer Indexing for amortizing indexing computation across layers, and Hierarchical Indexing for reducing the token-level scoring cost (\cref{sec:lsa}).

    \item Through extensive ablation studies, we show that the training-aware SI and CLI mechanisms improve both training and inference efficiency while maintaining performance parity with full attention. The training-free HI module provides further inference speedups with only marginal quality degradation (\cref{sec:exp,sec:efficiency}).

    \item We release LongCat-Flash-Lite-Sparse, an open-source model built with LSA, to facilitate research and practical deployment of sparse-attention architectures (\cref{sec:release_model}).
\end{itemize}

\section{Sparse Attention and the DSA Baseline}
\label{sec:dsa}

Before presenting our method, we review the evolution of sparse attention, formalize the DSA mechanism that serves as our baseline, and profile its efficiency bottlenecks.

\subsection{From Fixed Patterns to Retrieval-Based Sparse Attention}
\label{sec:frontier}

The central challenge in scaling self-attention~\citep{vaswani2017attention} to long contexts is the quadratic cost $\mathcal{O}(L^2)$ in sequence length $L$. Sparse attention methods address this by restricting each query to attend only a small subset of key-value positions. These methods differ primarily in \emph{how} they determine the positions attended by each query. Along this design dimension, approaches have evolved from content-blind fixed patterns to increasingly sophisticated learned retrieval mechanisms.

\textbf{Fixed patterns.} Early sparse attention methods define sparsity structures a priori, such as sliding windows, strided patterns, or designated global tokens~\citep{child2019sparse, beltagy2020longformer, zaheer2020bigbird}. These fixed patterns are hardware-friendly due to their regularity but entirely content-blind: the same positions are attended regardless of input.

\textbf{Query-aware but non-learnable retrieval.} A natural next step is to make the sparsity pattern content-dependent. Reformer~\citep{kitaev2020reformer} applies locality-sensitive hashing (LSH) to route queries and keys into shared buckets, restricting attention to within-bucket pairs. RetrievalAttention~\citep{liu2024retrievalattention} builds an approximate nearest-neighbor (ANN) index over stored keys and retrieves relevant KV entries at inference time. In both cases the selection is query-aware, yet the retrieval criterion cannot be directly optimized through training to improve selection quality.

\textbf{Coarse-grained learned retrieval.} The key leap in recent work is making the retrieval mechanism itself \emph{trainable}. MoBA~\citep{lu2025moba} and NSA~\citep{yuan2025nsa} partition the KV sequence into contiguous blocks and learn which blocks are most relevant to each query. Their indexing mechanisms directly reuse the core attention's query and key representations (after compression): MoBA computes block affinity via mean-pooled keys, while NSA repurposes its compression branch's attention scores as block importance signals. However, these block-level methods inherently blur token-level importance distinctions within each block, leading to possibly suboptimal performance relative to full attention on some long-context and reasoning tasks.

\textbf{Fine-grained learned retrieval.} DSA~\citep{deepseek2025dsa} advances to \emph{token-level} fine-grained learned sparse attention, scoring every individual token in the prefix for each query. Unlike MoBA and NSA which directly reuse the core attention's representations for scoring, DSA introduces a dedicated \emph{Lightning Indexer} with its own query and key projections, whose inputs are detached from the model's computation graph during training. This fine-grained indexing combined with the indexer's architectural independence enables DSA to achieve near-lossless performance relative to full attention. DSA has since been adopted by multiple production systems including DeepSeek-V3.2~\citep{deepseek2025v32, deepseek2025dsa} and GLM-5~\citep{zeng2026glm5}, validating its deployability at scale. 

\subsection{A Brief Recapitulation of DSA Mechanism}
\label{sec:dsa_recap}

We first review the formal details of DSA to establish notation used throughout this paper.

\paragraph{Indexer scoring.}
Given a query token with hidden state $\mathbf{h}_t$, the Lightning Indexer computes a saliency score for every token $s \leq t$:
\begin{equation}
\label{eq:indexer_score}
I_{t,s} = \sum_{j=1}^{H^I} w_{t,j}^I \cdot \mathrm{ReLU}\!\left(\mathbf{q}_{t,j}^I \cdot \mathbf{k}_s^I\right),
\end{equation}
where $H^I$ is the number of indexer heads, $\mathbf{q}_{t,j}^I$ and $w_{t,j}^I$ are derived from $\mathbf{h}_t$ via learned projections, and $\mathbf{k}_s^I$ is derived from $\mathbf{h}_s$. The indexer adopts an MQA~\citep{shazeer2019mqa} pattern: a single key $\mathbf{k}_s^I$ is shared across all indexer heads. The ReLU activation is chosen for throughput efficiency.

\paragraph{Top-$K$ selection and sparse attention.}
For each query position $t$, the indexer selects the $K$ highest-scoring tokens:
\begin{equation}
\label{eq:topk}
\mathcal{S}_t = \mathop{\mathrm{arg\,topK}} (\{I_{t,s}\}_{s \leq t}, \; K),
\end{equation}
and attention is computed only over this sparse subset:
\begin{equation}
\label{eq:sparse_attn}
\mathbf{u}_t = \mathrm{Attn}\!\left(\mathbf{h}_t,\; \{\mathbf{c}_s \mid s \in \mathcal{S}_t\}\right),
\end{equation}
where $\mathbf{c}_s$ denotes the MLA latent KV representation~\citep{liu2024deepseekv2} of token $s$. DSA is instantiated under MLA's absorbed MQA~\citep{shazeer2019mqa} mode, where each latent vector is shared across all query heads as the key-value entry. With $K = 2048$ in a 128K-token context, this achieves approximately 98.4\% sparsity while maintaining near-lossless quality.

\paragraph{Two-stage training.}
\label{sec:dsa_training}
DSA training proceeds from a pre-trained full-attention model in two stages:
\begin{enumerate}
\item \textbf{Dense warm-up}: The base model is frozen, and only the indexer parameters are trained. The objective aligns the indexer output with the aggregated attention distribution via KL divergence:
\begin{equation}
\label{eq:warmup_loss}
\mathcal{L}_I^{\text{dense}} = \sum_t D_{\mathrm{KL}}\!\left(\mathbf{p}_{t,:} \;\|\; \mathrm{Softmax}(\mathbf{I}_{t,:})\right),
\end{equation}
where $\mathbf{p}_{t,:}$ is the sum of attention weights across all heads, normalized to a distribution.

\item \textbf{Sparse training}: Both the indexer and the base model are trained jointly. The indexer is now supervised only over the selected token set $\mathcal{S}_t$, with the attention target renormalized over this subset:
\begin{equation}
\label{eq:sparse_loss}
\mathcal{L}_I^{\text{sparse}} = \sum_t D_{\mathrm{KL}}\!\left(\mathbf{p}_{t,\mathcal{S}_t} \;\big\|\; \mathrm{Softmax}\!\left(\mathbf{I}_{t,\mathcal{S}_t}\right)\right),
\end{equation}
where $\mathbf{p}_{t,\mathcal{S}_t}$ denotes the head-aggregated attention weights restricted to $\mathcal{S}_t$ and renormalized to a distribution, and $\mathbf{I}_{t,\mathcal{S}_t}$ the corresponding indexer scores. The indexer parameters are updated by this KL loss, while its input is detached from the model's computation graph so that the indexer's gradients do not perturb the base model. The base model parameters are updated solely via the next-token prediction loss.
\end{enumerate}

\paragraph{Why ``Lightning''.}
The indexer is designed to be lightweight relative to the core attention. It uses fewer heads (typically half that of the core attention) and a smaller head dimension (roughly one-third, since the indexer's nope dimension is much smaller than MLA's absorbed latent dimension). It further supports FP8 computation. These choices collectively minimize the per-token cost of indexing, justifying the ``Lightning'' designation.

\paragraph{Complexity.}
The overall computational cost of DSA consists of indexer scoring $\mathcal{O}(L^2)$ and sparse attention $\mathcal{O}(LK)$. Despite its low per-token cost, the indexer scales quadratically with $L$. Consequently, for sufficiently long sequences, the indexer becomes the dominant bottleneck and can incur greater computational cost than the core attention operation it is designed to serve. We quantify this crossover through systematic profiling in the next section.

\subsection{Profiling the Efficiency Bottlenecks of DSA}
\label{sec:profiling}

The execution cost of a DSA layer is partitioned between two primary operators whose latencies scale differentially with respect to the KV length $L$. For each incoming query token, the \emph{Lightning Indexer} (LI) must evaluate all $L$ prefix keys, resulting in a linear complexity ($\mathcal{O}(L)$). Conversely, the \emph{Sparse Flash Attention} (SFA) restricts its attention to a fixed set of $K$ selected tokens, thereby remaining essentially decoupled from $L$. This divergence creates a clear regime shift: SFA dominates computational overhead in short contexts, while LI takes over at extended sequence lengths, intersecting at approximately ${\sim}100$K tokens. This breakdown during the decoding stage is quantified in \cref{tab:decode_breakdown} for KV lengths ranging from 4K to 1024K. These regimes reveal two intrinsic efficiency bottlenecks in DSA’s sparse selection mechanism: \emph{Indexer Output Discontiguity} and \emph{Indexer High Overhead}. We examine both bottlenecks in detail below.

\begin{table}[ht]
\centering
\caption{Per-layer latency breakdown of baseline DSA during inference decode ($batch\_size=4$, $q_{\text{len}}=1$, BF16 precision, $K=2048$, causal masking). SFA is constant in $L$ (fixed $K$), while the indexer grows linearly, so the dominant cost shifts from SFA to LI near $\sim$100K.}
\label{tab:decode_breakdown}
\small
\begin{tabular}{ccccc}
\toprule
\textbf{KV Length} & \textbf{Indexer (ms)} & \textbf{SFA (ms)} & \textbf{Total (ms)} & \textbf{Indexer \%} \\
\midrule
4K   & 0.034 & 0.097 & 0.131 & 26\% \\
8K   & 0.036 & 0.096 & 0.132 & 27\% \\
16K  & 0.041 & 0.098 & 0.139 & 29\% \\
32K  & 0.053 & 0.098 & 0.151 & 35\% \\
64K  & 0.078 & 0.097 & 0.175 & 45\% \\
128K & 0.154 & 0.100 & 0.254 & 61\% \\
256K & 0.302 & 0.102 & 0.404 & 75\% \\
512K & 0.523 & 0.102 & 0.625 & 84\% \\
1024K & 0.930 & 0.102 & 1.032 & \textbf{90\%} \\
\bottomrule
\end{tabular}
\end{table}

\paragraph{Bottleneck 1: Non-coalesced memory access arising from Indexer Output Discontiguity.}
Token-level sparse selection forces each memory transaction to retrieve a single, non-contiguous KV vector, severely degrading High Bandwidth Memory (HBM) utilization. On our AI accelerators, a single core can sustain approximately $50$ in-flight cachelines ($512$\,B each) under ideal memory coalescing, maintaining a memory window of ${\sim}25.6$\,KB. In DSA, however, each selected token is fetched via an independent gather of a single latent KV vector ($1{,}152$\,B in BF16), which spans only $3$ cachelines. Consequently, two inefficiencies compound: first, this gather occupies merely $3$ of the ${\sim}50$ outstanding memory slots (yielding $\approx 6\%$ memory-level parallelism); second, even within these $3$ cachelines, data packing efficiency is only $\approx 75\%$. The resulting net effective bandwidth thus plummets to a mere $\approx \mathbf{4.5\%}$ ($\approx 1/22$) of peak performance.

While the aforementioned analysis focuses on the forward pass during inference, this output discontiguity inflicts a more severe performance penalty during training. In the backward pass, gradient updates rely on \texttt{scatter\_add} operations over the identical non-contiguous token indices. Due to the scattered nature and input-dependent selection of these indices, independent computing cores frequently attempt to write to overlapping KV gradient regions in HBM. When different cores concurrently target identical gradient positions, these probabilistic write conflicts force the HBM transactions to serialize, severely degrading hardware concurrency. The indexer backward pass suffers from this same cross-core serialization as it scatters gradients across all $L$ scored positions.

\paragraph{Bottleneck 2: High Indexer Overhead from linear-scaling scoring and Top-$K$ selection.}

Consider long-context decoding, where a fixed number of query tokens attends to a KV cache whose length $L$ grows continuously. The computational cost of the LI scales linearly with $L$ for each query because both of its internal stages operate over the entire context. Specifically, the scoring stage performs an MQA-style QK matrix multiplication with $\mathcal{O}(L)$ complexity, after which Top-$K$ selection incurs an additional $\mathcal{O}(L)$ cost over all candidate scores. By contrast, SFA attends to a fixed budget of $K$ selected tokens, making its decoding cost independent of $L$.

This asymmetry explains the latency breakdown reported in \cref{tab:decode_breakdown}. As the context length increases from 4K to 1024K, SFA latency remains nearly constant at ${\sim}0.10$\,ms, whereas LI latency increases by $27\times$, from $0.034$\,ms to $0.930$\,ms. The dominant per-layer bottleneck therefore shifts from SFA at short context lengths to LI at long context lengths. At a 1024K context length, the indexer accounts for up to $90\%$ of the total per-layer latency. Moreover, although LI has linear complexity per decoding query, applying it to every query during prefill or training results in an aggregate $\mathcal{O}(L^2)$ indexing cost, fundamentally limiting the scalability of standard DSA.

\section{LongCat Sparse Attention (LSA)}
\label{sec:lsa}

Building upon the DSA baseline formalized in \cref{sec:dsa_recap}, LSA introduces three orthogonal efficiency improvements to the indexer: Streaming-Aware Indexing (SI; \cref{sec:streaming}), Cross-Layer Indexing (CLI; \cref{sec:cli}), and Hierarchical Indexing (HI; \cref{sec:hierarchical}). Each module targets a distinct bottleneck dimension identified in \cref{sec:profiling} and can compose naturally with the others. We describe each component in detail in the remainder of this section.

\subsection{Streaming-Aware Indexing: Improving Locality for Hardware-Aligned Coalesced Access}
\label{sec:streaming}

\paragraph{Attention sinks and streaming heads.}
StreamingLLM~\citep{xiao2023streamingllm} reveals that a small number of initial tokens act as attention sinks and absorb a disproportionate amount of attention weight because of softmax normalization constraints. It further demonstrates that retaining only these sink tokens together with a local sliding window is sufficient to maintain stable perplexity during streaming inference. DuoAttention~\citep{xiao2024duoattention} further deepens this finding by showing that attention heads exhibit clear functional specialization: \emph{Streaming Heads} primarily attend to sinks and recent tokens (capturing local context and numerical stability), while \emph{Retrieval Heads} are responsible for long-range information retrieval.

\begin{figure}[t]
    \centering
    \includegraphics[width=0.85\linewidth]{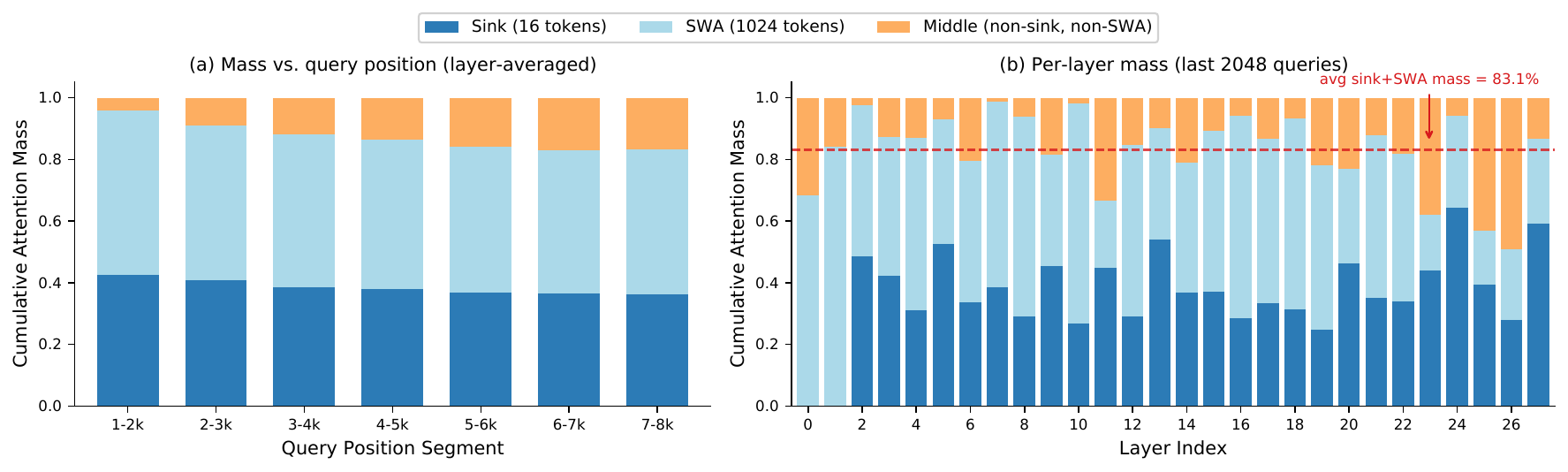}
    \caption{Attention mass distribution on a full-attention \longcat (69B-A3B) model with 14 short-cut MoE blocks, each containing two serial attention operations (28 attention layers total). Data: 20 InfBench-QA samples, seq\_len = 8192. \textbf{(a)} Layer-averaged breakdown by query position segment: the streaming ratio converges to $\sim$83\% beyond 5k tokens, confirming the pattern persists at long range. \textbf{(b)} Per-layer breakdown over the last 2048 queries: the streaming regions (Sink + SWA) capture an average of $\sim$83\% of attention mass across all layers.}
    \label{fig:streaming_mass}
\end{figure}

\paragraph{Streaming-aware budget partitioning.}
Departing from head-wise classification, we analyze the aggregate attention mass distribution across all heads. Our empirical analysis reveals a persistent \emph{streaming pattern}: the sink and sliding window regions consistently capture a substantial share of the attention weight (see \cref{fig:streaming_mass}). Recognizing this as a stable structural feature, we formalize their allocation as a deterministic budget. This approach allows us to partition the attention operation into a hardware-aligned fixed streaming component (sink and sliding window) and a dynamic sparse component. Formally, we decompose the attended token set into three disjoint subsets:
\begin{equation}
\label{eq:set_split}
\mathcal{S}_t \;=\; \underbrace{\mathcal{S}_{\text{sink}} \;\cup\; \mathcal{S}_{\text{swa}}}_{\text{fixed streaming budgets}} \cup\, \mathcal{S}_{\text{sparse}},
\end{equation}
where $K_{\text{sink}} = |\mathcal{S}_{\text{sink}}|$, $K_{\text{swa}} = |\mathcal{S}_{\text{swa}}|$, and $K_{\text{sparse}} = |\mathcal{S}_{\text{sparse}}|$ denote the sizes of each subset, satisfying $\kmax = K_{\text{sink}} + K_{\text{swa}} + K_{\text{sparse}}$. Specifically, $\mathcal{S}_{\text{sink}} = \{1, \ldots, K_{\text{sink}}\}$ covers the sink region, $\mathcal{S}_{\text{swa}} = \{t - K_{\text{swa}} + 1, \ldots, t\}$ is a sliding window around the query, and $\mathcal{S}_{\text{sparse}}$ contains the remaining $K_{\text{sparse}}$ tokens dynamically selected by the indexer from the non-fixed positions. The indexer's scoring range correspondingly shrinks:
\begin{equation}
\label{eq:streaming_indexer}
\mathcal{S}_{\text{sparse}} = \mathop{\mathrm{arg\,topK}} (\{I_{t,s}\}_{s \notin \mathcal{S}_{\text{sink}} \cup \mathcal{S}_{\text{swa}}}, \; K_{\text{sparse}}),
\end{equation}

\paragraph{Training.}
Under the standard two-stage training (\cref{sec:dsa_training}), the dense warm-up stage requires no modification: the indexer is supervised against the full-sequence attention distribution (\cref{eq:warmup_loss}). In the sparse training stage, the distillation target becomes the attention distribution renormalized over the selected subset $\mathcal{S}_t$. Although at inference time the indexer only scores tokens in the middle region ($s \notin \mathcal{S}_{\text{sink}} \cup \mathcal{S}_{\text{swa}}$), during training we deliberately distill the indexer over the \emph{entire} selected set, including the sink and SWA portions. Since these streaming regions capture a substantial fraction of the attention mass (\cref{fig:streaming_mass}), including them in the distillation target provides richer supervisory signal and enables the indexer to more thoroughly learn the overall attention structure, improving its ability to identify the most relevant tokens in the remaining middle region.

\paragraph{Design choices and benefits.}
We set sink size $K_{\text{sink}} = 16$ and sliding window size $K_{\text{swa}} = 1024$, yielding a fixed-to-sparse budget ratio of approximately 1:1 (determined by ablation, see \cref{sec:ablation}), which makes roughly 50\% of the selected tokens reside in contiguous memory regions. This partitioning yields three benefits: (a)~the sink and sliding-window portions are accessed as contiguous blocks, enabling efficient coalesced HBM reads; (b)~the indexer's effective scoring range shrinks from $L$ to $L - K_{\text{sink}} - K_{\text{swa}}$, modestly reducing computation; and (c)~the deterministic structure provides a natural interface for KV cache offloading and speculative decoding, where consecutive decode steps share predictable cache regions.

\subsection{Cross-Layer Indexing: Amortizing Indexing Overhead via Inter-layer Pattern Redundancy}
\label{sec:cli}

\paragraph{Stability of salient tokens across layers.}
Recent studies~\citep{yang2024tidaldecode, deshmukh2025kascade, gao2026hysparse} have shown that the set of salient tokens, defined as those receiving the highest attention mass, remains remarkably stable across consecutive Transformer layers. This observation has motivated a family of sparse attention methods that exploit cross-layer stability to avoid repeatedly selecting salient tokens in each layer. These methods typically designate a few full-attention layers as oracles to identify important tokens and let remaining sparse layers reuse their selections. A natural question arises: does DSA's Lightning Indexer at a given layer have the capacity to identify the shared salient tokens across itself and subsequent layers?

\begin{figure}[t]
    \centering
    \includegraphics[width=0.95\linewidth]{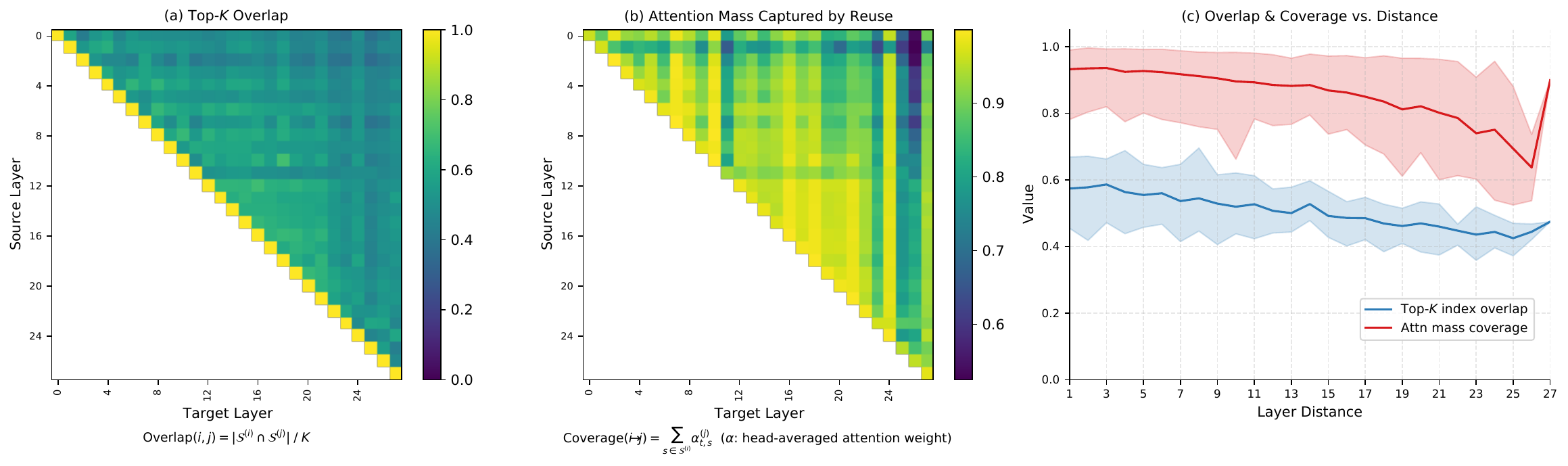}
    \caption{Cross-layer Top-$\kmax$ index analysis on full-attention \longcat (69B-A3B). Each layer independently selects its Top-$\kmax$ tokens from the full attention distribution. Only the upper triangle is shown in (a) and (b), corresponding to the CLI reuse direction, where an earlier source layer provides the index set for later target layers). \textbf{(a)} Pairwise overlap of Top-$\kmax$ token sets. \textbf{(b)} Cumulative attention mass captured when reusing one layer's Top-$\kmax$ on another. \textbf{(c)} Mean overlap and coverage as a function of layer distance, with shaded regions indicating min--max range across all layer pairs at each distance; adjacent layers share $\sim$57\% of their Top-$\kmax$ budget while retaining $\sim$93\% of attention mass. Statistics are collected from 20 InfBench-QA samples with sequence length 8192, computed over the last 2048 query positions of each sample..}
    \label{fig:cli_overlap}
\end{figure}

We verify this on \longcat by letting each layer independently run its own indexer and measuring cross-layer agreement (\cref{fig:cli_overlap}). Adjacent layers share 57.4\% of their Top-$\kmax$ budget on average. Despite this partial overlap, reusing an adjacent layer's index set still captures 93.2\% of the target layer's attention mass. This cross-layer consistency opens up the possibility of reusing indexer outputs across layers.

\paragraph{Enabling cross-layer indexer reuse with cross-layer distillation.}
We partition consecutive layers into CLI groups of size $N$. Only the first layer in each group executes the indexer; the subsequent $N - 1$ layers directly reuse the index set $\mathcal{S}_t$ produced by the first layer, reducing the total number of indexing passes from $L_{\text{layers}}$ to $L_{\text{layers}} / N$. Naively reusing indices without training adaptation degrades performance (\cref{sec:ablation}), because each indexer is originally trained to predict saliency only for its own layer. To enable faithful cross-layer reuse, we make a minimal modification to the standard DSA distillation loss (\cref{eq:warmup_loss}): instead of supervising each indexer against only its own layer's attention, the indexer at the first layer of each group is trained to predict the attention patterns of \emph{all} layers within the group:
\begin{equation}
\label{eq:cli_loss}
\mathcal{L}_{\text{CLI}} = \sum_{i=0}^{N-1} \mathcal{L}_I^{(l+i)},
\end{equation}
where $l$ is the first layer in the group, $N$ is the group size, and $\mathcal{L}_I^{(l+i)}$ is the indexer distillation loss that supervises the shared indexer output against the attention distribution of layer $l+i$. Note that this cross-layer distillation applies to both the dense warm-up and sparse training stages. At inference time, the shared indexer runs once and broadcasts the resulting index set to all layers in the group.

\paragraph{Integration with Multi-Token Prediction.}
\label{sec:mtp_integration}
Multi-Token Prediction (MTP)~\citep{liu2024deepseekv3, zeng2026glm5} appends $D$ sequential prediction steps (each a Transformer layer with its own attention) after the main model to predict $D$ additional future tokens, where each step combines the previous step's representation with the next token's embedding. We extend CLI to the MTP dimension: all $D$ MTP steps form their own independent CLI group (separate from the main model's groups) and share a single index set produced by the first MTP step's indexer, trained via cross-layer distillation: $\mathcal{L}_{\text{CLI}}^{\text{MTP}} = \sum_{k=1}^{D} \mathcal{L}_I^{(\mathrm{MTP}_k)}$. Unlike CLI in the main model, where adjacent layers process the same input and naturally exhibit correlated saliency patterns, MTP steps operate sequentially. Each step attends to representations that incorporate different future token embeddings, which may make index reuse across MTP steps appear less justified. However, the cross-layer distillation loss ensures that the shared indexer produces a selection jointly optimized for all MTP steps. At inference time, the first MTP step's indexer has already been trained to account for the attention patterns of subsequent steps, making CLI reuse across MTP steps equally well-founded. Ablation experiments (\cref{sec:ablation}) confirm that this approach maintains quality parity with independent per-step indexing.

\paragraph{Design choices.}
Although our analysis reveals that certain layers can sustain high attention mass coverage even with large reuse depths (\cref{fig:cli_overlap}c), practical LLM architecture design favors simplicity and uniform structure for robust scaling across model sizes. We therefore adopt a uniform interleave pattern with a fixed group size $N$ across all layers. We further constrain $N$ to be even for two reasons: (1)~the \longcat series adopts a shortcut-connected architecture~\citep{cai2024shortcut,team2025longcat} where each shortcut layer contains two serial attention operations, and (2)~even group sizes ensure uniform partitioning across pipeline-parallel stages during training. From \cref{fig:cli_overlap}(c), we observe that the minimum coverage across layer pairs begins to drop noticeably when the reuse distance exceeds 4. Ablation experiments (\cref{sec:ablation}) show that $N = 4$ incurs measurable accuracy loss on long-context validation, so we set $N = 2$, which halves the indexing compute with no measurable quality loss. For MTP steps, since they serve as draft proposers for speculative decoding whose outputs are verified by the main model, their accuracy does not affect final generation quality. We therefore let all 3 MTP steps share a single index (i.e., $N = 3$) to maximize the efficiency gain.

\paragraph{Relation to concurrent work.}
After our CLI strategy was validated and deployed in production, we became aware of IndexCache~\citep{bai2026indexcache}, an independent and concurrent effort that arrives at a highly similar training-aware cross-layer distillation mechanism for index reuse. 
Despite the shared core idea, several differences are worth noting.
(1)~\emph{Architecture}: we validate on \longcat's shortcut-connected structure (two serial attention operations per shortcut layer), whereas IndexCache targets a standard Transformer layout.
(2)~\emph{Effective reuse ratio}: IndexCache reports that 1/4 indexer retention (i.e., $N{=}4$) remains within 0.4\% of baseline on their evaluation benchmarks, whereas our ablations (\cref{sec:ablation}) show that $N{=}4$ incurs measurable accuracy loss on long-context validation, leading us to adopt the more conservative $N{=}2$. This discrepancy might stem from the differences in model architecture and context length noted above.
(3)~\emph{Composability}: beyond validating CLI in isolation, we demonstrate that it composes effectively with the other two LSA components (Streaming-aware and Hierarchical Indexing), achieving greater aggregate speedup than any single strategy alone.
(4)~\emph{Multi-Token Prediction}: we further verify that CLI remains effective across MTP steps, extending index reuse to the speculative decoding path.

\subsection{Hierarchical Indexing: Coarse-to-Fine Sparse Selection}
\label{sec:hierarchical}

The indexer scores every token in the non-fixed region for each query, yielding $\mathcal{O}(L)$ per-query cost that dominates latency at long contexts. Can we eliminate most irrelevant tokens cheaply before invoking fine-grained scoring? Our solution is a two-stage coarse-to-fine scoring scheme: a cheap coarse stage first recalls a small subset of candidate sub-sequences, shrinking the candidate set on which the expensive fine-grained scoring is then performed.

\paragraph{Stage 1: Block-level coarse filtering.}

We partition the sequence into contiguous pages of size $P$, and first perform a coarse selection to obtain Top-$M$ pages. For each page, we split it into sub-blocks of $B$ tokens each, and precompute the per-dimension mean of each sub-block's constituent keys: $\mathbf{k}_{n}^{\mean} = \mean_{s \in \text{sub-block}_n} \mathbf{k}_{s}^I$. The coarse saliency score for every page $p \leq \lceil t/P \rceil $ is computed by aggregating its sub-block scores in a manner similar to \cref{eq:indexer_score}:
\begin{equation}
\label{eq:block_score}
I_{t,p}^{\text{page}} = \sum_{j=1}^{H^I} w_{t,j}^I \cdot \sum_{n \in \text{page}_p} \mathrm{ReLU}\!\left(\mathbf{q}_{t,j}^I \cdot \mathbf{k}_{n}^{\mean}\right),
\end{equation}
where $H^I$ is the number of indexer heads and $\mathbf{q}_{t,j}^I$, $w_{t,j}^I$ are the same per-head query and weight as in \cref{eq:indexer_score}. We select the Top-$M$ pages as candidates according to their saliency scores, and let $\mathcal{S}_t^{\text{page}}$ denote the set of tokens they cover, reducing the scoring space from $\mathcal{O}(L)$ tokens to $M \cdot P$:
\begin{equation}
\label{eq:page_topM}
\mathcal{S}_{t}^{\text{page}} = \bigcup_{p \in \mathcal{P}_t} \bigcup_{n \in \text{page}_p} \text{sub-block}_n, ~ \text{where} ~ \mathcal{P}_t  = \mathop{\mathrm{arg\,topK}}\left(\{I_{t,p}^{\text{page}}\}_{p \leq \lceil t/P \rceil},\; M\right).
\end{equation}

\paragraph{Stage 2: Token-level refinement.}
For the $M$ selected candidate pages (comprising $M \cdot P$ tokens), we apply the standard indexer scoring of \cref{eq:indexer_score} to obtain per-token scores. Specifically, using the same indexer query $\mathbf{q}_{t,j}^I$, the fine-grained indexer score for every token $s$ within the recalled pages is computed as
\begin{equation}
\label{eq:fine_score}
I_{t,s} = \sum_{j=1}^{H^I} w_{t,j}^I \cdot \mathrm{ReLU}\!\left(\mathbf{q}_{t,j}^I \cdot \mathbf{k}_s^I\right), \qquad s \in \mathcal{S}_t^{\text{page}}.
\end{equation}

We select the final Top-$K_{\text{sparse}}$ tokens, yielding the same sparse set $\mathcal{S}_{\text{sparse}}$ of \cref{eq:streaming_indexer} but scored only over the recalled pages:
\begin{equation}
\label{eq:hierarchical_complexity}
\mathcal{S}_{\text{sparse}} = \mathop{\mathrm{arg\,topK}}\left(\{I_{t,s}\}_{s \in \mathcal{S}_t^{\text{page}}\cap [1, t]},\; K_{\text{sparse}}\right).
\end{equation}
As analyzed in \cref{sec:hi_op}, the Top-$K$ selection dominates the indexer cost while the scoring itself is comparatively cheap, so we account only for the selection cost. Under this measure, the two Top-K operations in \cref{eq:page_topM,eq:hierarchical_complexity} cost $\mathcal{O}(L/P)$ over pages and $\mathcal{O}(M \cdot P)$ over the recalled tokens, respectively, reducing the original $\mathcal{O}(L)$ per-query selection to $\mathcal{O}(L/P + M \cdot P)$.

\paragraph{Design choices.}
Hierarchical Indexing is training-free: it requires no additional parameters or fine-tuning, operating purely as an inference-time optimization over the existing indexer. 
The mean value of the blocks is precomputed once per sequence and stored in a cache. 

We fix the page size $P$ to 128, while treating the sub-block size $B$ and candidate count $M$ as tunable hyperparameters. The ablation study in \cref{sec:abl_hier} identifies $B{=}8$ and $M{=}1024$ as the optimal configuration. Profiling under this setting (\cref{tab:hi_infer}) reveals that the two-stage design introduces non-negligible overhead from block-mean maintenance, coarse-grained scoring, and candidate gathering. As a result, Hierarchical Indexing provides net efficiency gains only for sequence lengths of $\ge256$K, and we enable it adaptively beyond this crossover point.

\section{Kernel Design and Efficiency Analysis}
\label{sec:efficiency}

\subsection{Efficient Kernel Design and Implementation}
\label{sec:kernel}

Among the three components of LSA, CLI reuses the indexer across layers and therefore requires no kernel-level modification. This section thus focuses on the kernel design of SI and HI, which reshape the DSA layer at the operator level to address the bottlenecks identified in \cref{sec:profiling}.

\subsubsection{Hybrid Sparse Attention Operator for SI: Partitioning and Overlapped Execution}
Recall that SI partitions the attention budget into two components of approximately equal size---a sliding window $\mathcal{S}_{\text{swa}}$ and a dynamic sparse selection $\mathcal{S}_{\text{sparse}}$ (plus a small sink $\mathcal{S}_{\text{sink}}$ of only $16$ tokens, which we omit in the analysis below for brevity). To translate this algorithmic partitioning into tangible hardware efficiency gains, we design an integrated operator, Hybrid Sparse Attention (HFA).

In the forward pass, instead of processing the entire attention budget with a single SFA operator, HFA decomposes core attention into an SFA operator over $\mathcal{S}_{\text{sparse}}$ and an SWA operator over $\mathcal{S}_{\text{swa}}$. The two operators are dispatched to separate non-blocking hardware streams for overlapped execution, and their partial outputs are merged through online-softmax rescaling.

In the backward pass, beyond confronting non-coalesced memory access issues similar to those in the forward pass, HFA encounters a unique and more severe bottleneck during gradient accumulation. Specifically, gradient updates in the SFA backward pass are performed by allocating a gradient buffer for the entire sequence length in High Bandwidth Memory (HBM) and executing a \texttt{scatter\_add} operation over the selected sparse index set. This process constitutes the dominant execution cost of the backward pass. Beyond the low memory bandwidth utilization caused by discontinuous reads, it introduces a severe write-conflict bottleneck: when different processing cores select overlapping indices, they must write to the same physical memory addresses in HBM, resulting in serialized writes to shared cachelines that heavily restrict throughput. By assigning roughly half of the fixed attention budget to the sliding window $\mathcal{S}_{\text{swa}}$, HFA correspondingly reduces the active sparse index set, thereby decreasing the number of discrete gather/scatter operations and lowering the probability of write conflicts during gradient accumulation.

\begin{table}[ht]
\centering
\caption{Core attention latency (ms) under a representative training setting. Measurements use $batch\_size = 1$, $q_{\text{len}} = 8192$, BF16, and $K=2048$. For SI, we use $K_{\text{sparse}} = 1024$ and $K_{\text{swa}} = 1024$, omitting $K_{\text{sink}} = 16$ for simplicity. SFA uses the full budget for dynamic sparse selection, while HFA splits it into parallel sparse (SFA) and contiguous (SWA) branches, with outputs merged via online-softmax rescaling.
}
\label{tab:si_train}
\small
\begin{tabular}{llcccccccc}
\toprule
& \textbf{KV Length} & 8K & 16K & 32K & 64K & 128K & 256K & 512K & 1024K \\
\midrule
\multirow{3}{*}{\rotatebox{90}{\textbf{fwd}}}
& $\text{SFA}$          & 23.59 & 23.65 & 23.64 & 23.71 & 23.98 & 24.25 & 24.46 & 36.80 \\
& $\text{HFA}$        & 18.46 & 19.31 & 19.70 & 19.15 & 19.53 & 19.42 & 19.84 & 19.27 \\
& \emph{speedup}     & 1.28$\times$ & 1.22$\times$ & 1.20$\times$ & 1.24$\times$ & 1.23$\times$ & 1.25$\times$ & 1.23$\times$ & \textbf{1.91$\times$} \\
\midrule
\multirow{3}{*}{\rotatebox{90}{\textbf{bwd}}}
& $\text{SFA}$         & 165.99 & 215.42 & 238.60 & 267.57 & 277.92 & 296.01 & 319.91 & 354.48 \\
& $\text{HFA}$       & 113.86 & 135.16 & 137.77 & 155.02 & 165.74 & 175.04 & 189.85 & 220.12 \\
& \emph{speedup}     & 1.46$\times$ & 1.59$\times$ & \textbf{1.73$\times$} & \textbf{1.73$\times$} & 1.68$\times$ & 1.69$\times$ & 1.69$\times$ & 1.61$\times$ \\
\bottomrule
\end{tabular}
\end{table}

\begin{table}[ht]
\centering
\caption{Core attention and full-layer latency (ms) under representative inference settings. Results are measured with BF16 precision and $K = 2048$ for prefill ($batch\_size = 1$, $q_{\text{len}} = 2048$) and decode ($batch\_size = 4$, $q_{\text{len}} = 1$). We compare core attention latency (SFA vs. HFA) and full-layer latency including indexer overhead: SFA + LI (Lightning Indexer) versus HFA + SI (Streaming-Aware Indexing).}
\label{tab:si_infer}
\scriptsize
\begin{tabular}{llccccccccc}
\toprule
& \textbf{KV Length} & 4K & 8K & 16K & 32K & 64K & 128K & 256K & 512K & 1024K \\
\midrule
\multirow{6}{*}{\rotatebox{90}{\textbf{Prefill}}}
& $\text{SFA}$            & 5.946 & 5.961 & 5.936 & 5.964 & 6.004 & 6.017 & 6.099 & 6.672 & 9.737 \\
& $\text{HFA}$             & 3.775 & 3.797 & 3.809 & 3.821 & 3.826 & 3.863 & 3.857 & 4.170 & 5.756 \\
& \emph{speedup}      & 1.57$\times$ & 1.57$\times$ & 1.56$\times$ & 1.56$\times$ & 1.57$\times$ & 1.56$\times$ & 1.58$\times$ & 1.60$\times$ & \textbf{1.69$\times$} \\
\cmidrule(l){2-11}
& $\text{SFA} + \text{LI}$          & 6.644 & 7.401 & 8.877 & 11.902 & 17.955 & 30.006 & 54.137 & 102.848 & 202.180 \\
& $\text{HFA} + \text{SI}$           & 4.471 & 5.242 & 6.763 & 9.787 & 15.825 & 27.907 & 51.829 & 100.479 & 198.407 \\
& \emph{speedup}      & \textbf{1.49$\times$} & 1.41$\times$ & 1.31$\times$ & 1.22$\times$ & 1.13$\times$ & 1.08$\times$ & 1.04$\times$ & 1.02$\times$ & 1.02$\times$ \\
\midrule
\multirow{6}{*}{\rotatebox{90}{\textbf{Decode}}}
& $\text{SFA}$             & 0.097 & 0.096 & 0.098 & 0.098 & 0.097 & 0.100 & 0.102 & 0.102 & 0.102 \\
& $\text{HFA}$              & 0.085 & 0.083 & 0.085 & 0.085 & 0.087 & 0.081 & 0.083 & 0.084 & 0.081 \\
& \emph{speedup}      & 1.14$\times$ & 1.16$\times$ & 1.15$\times$ & 1.15$\times$ & 1.11$\times$ & 1.23$\times$ & 1.24$\times$ & 1.21$\times$ & \textbf{1.26$\times$} \\
\cmidrule(l){2-11}
& $\text{SFA} + \text{LI}$          & 0.131 & 0.132 & 0.139 & 0.151 & 0.175 & 0.254 & 0.404 & 0.625 & 1.032 \\
& $\text{HFA} + \text{SI}$           & 0.116 & 0.116 & 0.124 & 0.135 & 0.164 & 0.224 & 0.374 & 0.588 & 0.995 \\
& \emph{speedup}      & 1.13$\times$ & \textbf{1.14$\times$} & 1.12$\times$ & 1.12$\times$ & 1.07$\times$ & 1.13$\times$ & 1.08$\times$ & 1.06$\times$ & 1.04$\times$ \\
\bottomrule
\end{tabular}
\end{table}

\cref{tab:si_train,tab:si_infer} summarize the speedups achieved by our optimized HFA over the SFA baseline. Under the training configuration (\cref{tab:si_train}), HFA accelerates core attention by up to $1.91\times$ in the forward pass and $1.73\times$ in the backward pass. For inference (\cref{tab:si_infer}), we evaluate improvements at both the core-attention and full-layer levels. At the core-attention level, HFA achieves consistent speedups of $1.56$--$1.69\times$ during prefill and $1.11$--$1.26\times$ during decode. After including indexer overhead, however, the full-layer speedup becomes increasingly constrained by indexing cost, reaching a peak of $1.49\times$ for prefill at 4K context and $1.14\times$ for decode at 8K context, before gradually diminishing to $1.02$--$1.04\times$ at longer contexts. This result motivates the complementary roles of our designs: SI targets short-context efficiency by alleviating core-attention bottlenecks, while CLI and HI address long-context scenarios where indexing overhead dominates.

\subsubsection{Index-Selection Operator for HI: Two-Stage Coarse-to-Fine Selection}
\label{sec:hi_op}

\begin{table}[b]
\centering
\caption{Indexer latency (ms) under representative prefill settings ($q_{\text{len}} = 2048$, BF16). We compare the proposed hierarchical two-stage indexer with the flat baseline. The hierarchical design uses $B = 8$, $P = 128$, and $M = 1024$ pages in Stage 1 ($128\text{k}$-token candidate budget), followed by Stage 2 refinement to select the final $K_{\text{sparse}}$ tokens. Speedup is computed as Flat / Total.}
\label{tab:hi_infer}
\small
\begin{tabular}{lcccccc}
\toprule
\textbf{KV Length} & 32K & 64K & 128K & 256K & 512K & 1024K \\
\midrule
HI Stage 1          & 0.666 & 1.264 & 2.457 & 4.912 & 9.758 & 19.162 \\
HI Stage 2          & 6.814 & 13.353 & 27.769 & 27.769 & 27.769 & 27.769 \\
HI Total            & 7.480 & 14.617 & 30.226 & 32.681 & 37.527 & 46.931 \\
\midrule
Flat LI             & 5.934 & 11.950 & 23.977 & 48.025 & 96.139 & 192.698 \\
\emph{speedup}      & 0.79$\times$ & 0.82$\times$ & 0.79$\times$ & 1.47$\times$ & 2.56$\times$ & \textbf{4.11$\times$} \\
\bottomrule
\end{tabular}
\end{table}

Within the indexing computation, the main bottleneck is not score computation (\cref{eq:indexer_score}) but Top-$K$ selection (\cref{eq:topk}). Score computation benefits from high-throughput matrix processing units, whereas Top-$K$ requires sorting the full candidate set on much slower vector processing units. To overcome the latency bottleneck caused by global Top-$K$ selection in long-context regimes, we implement a custom operator for the Hierarchical Indexer (HI) formulated in \cref{sec:hierarchical}. We then profile the execution latency of this proposed two-stage indexer (Stage~1 coarse page recall followed by Stage~2 fine-grained token selection) against the flat Lightning Indexer (LI) baseline in \cref{tab:hi_infer}. 

Our analysis reveals two distinct performance regimes. Below the recall budget ($L{\le}128$K), Stage~2 still processes nearly the entire sequence and its latency scales with $L$. Consequently, the additional overhead of the two-stage design outweighs its benefits, resulting in a net slowdown of $0.79$--$0.82\times$. We therefore enable HI only beyond the crossover point (around $200$K tokens in this setting), while falling back to the flat LI for shorter contexts. Beyond the recall budget, Stage~2 saturates at a constant latency of $27.8$ms, whereas the flat baseline continues to grow linearly with context length. As a result, the advantage of HI increases substantially with longer contexts, achieving $4.11\times$ speedups at 1024K tokens.

\subsection{Attention Layer Training Speedup}
\label{sec:training_speedup}
We compare the training efficiency of LSA and DSA by measuring the forward and backward latency of a single attention layer across a spectrum of context lengths, reflecting practical production training setups.

\begin{figure}[t]
    \centering
    \includegraphics[width=\linewidth]{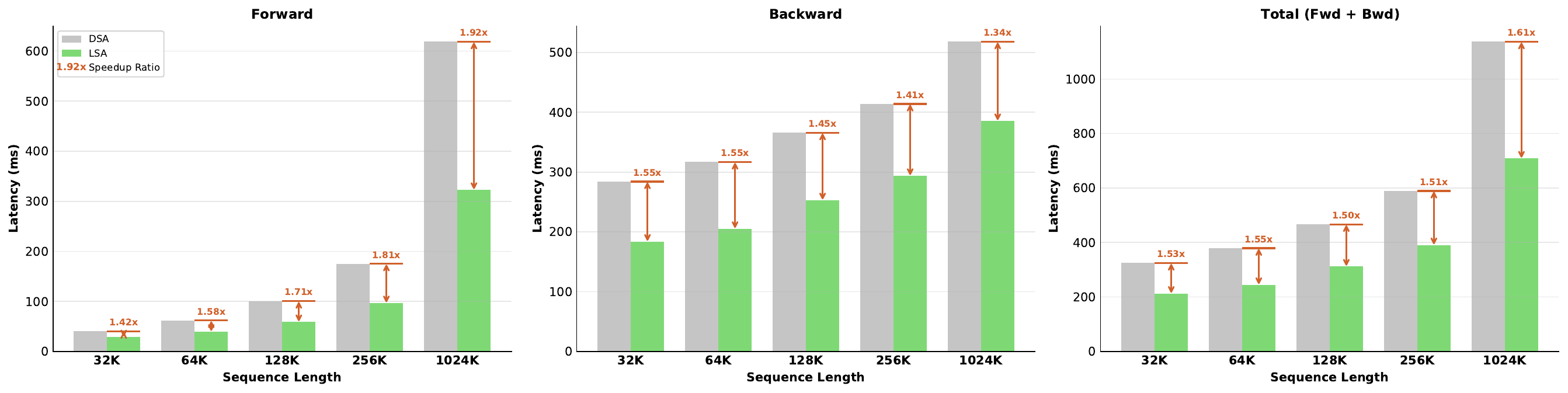}
    \caption{Single-attention-layer training latency of LSA and DSA across context lengths. Bars report forward, backward, and total latency, including kernel execution and CP communication; arrows indicate LSA speedup over DSA. The CP degrees are 4, 8, 16, 32, and 128 for context lengths of 32K, 64K, 128K, 256K, and 1024K, respectively.}
    \label{fig:train_layer}
\end{figure}

\paragraph{Speedup analysis.}
As shown in \cref{fig:train_layer}, LSA consistently reduces total attention-layer latency, achieving a $1.53\times$ speedup at 32K and a $1.61\times$ speedup at 1024K. Since HI is used only for inference, the training gains come from SI and CLI:
(1) \emph{CLI reduces forward-only indexer overhead.} By training a shared indexer across $N=2$ layers through cross-layer distillation, CLI amortizes one indexer forward over two attention layers. Since this sharing does not reduce backward computation, CLI exclusively benefits the forward pass, with larger gains emerging at long contexts where indexer cost dominates.
(2) \emph{SI accelerates both passes, especially backward.} As shown in \cref{sec:kernel}, SI improves both forward and backward efficiency by resolving the SFA backward write conflict. Since the backward pass is intrinsically more expensive than the forward pass, eliminating this bottleneck translates into a larger absolute saving and a stronger contribution to overall training throughput.
Together, they deliver $1.42$--$1.92\times$ forward speedup and $1.34$--$1.55\times$ backward speedup.

\subsection{End-to-End Inference Speedup}
\label{sec:inference_speedup}

We compare the end-to-end inference efficiency of LSA and DSA at the \longcat (69B-A3B) model scale under representative prefill and decoding workloads.

\paragraph{End-to-end performance.}
To support long-context inference, we use KV-cache Partition (KVP) to shard the KV cache across multiple ranks, alleviating per-device memory pressure. KVP is enabled for decoding at sequence lengths of at least 256K; detailed serving configurations and implementation are provided in \cref{app:serving_setup}.
We enable SI and CLI for all sequence lengths. HI is enabled only for prefill at sequence lengths of at least 256K, following the crossover point in \cref{tab:hi_infer}. We disable HI during decoding because its two-stage overhead outweighs its benefit at short context lengths, while KVP reduces the per-rank KV length in the long-context regime where HI would otherwise be beneficial.

\begin{figure}[t]
    \centering
    \includegraphics[width=\linewidth]{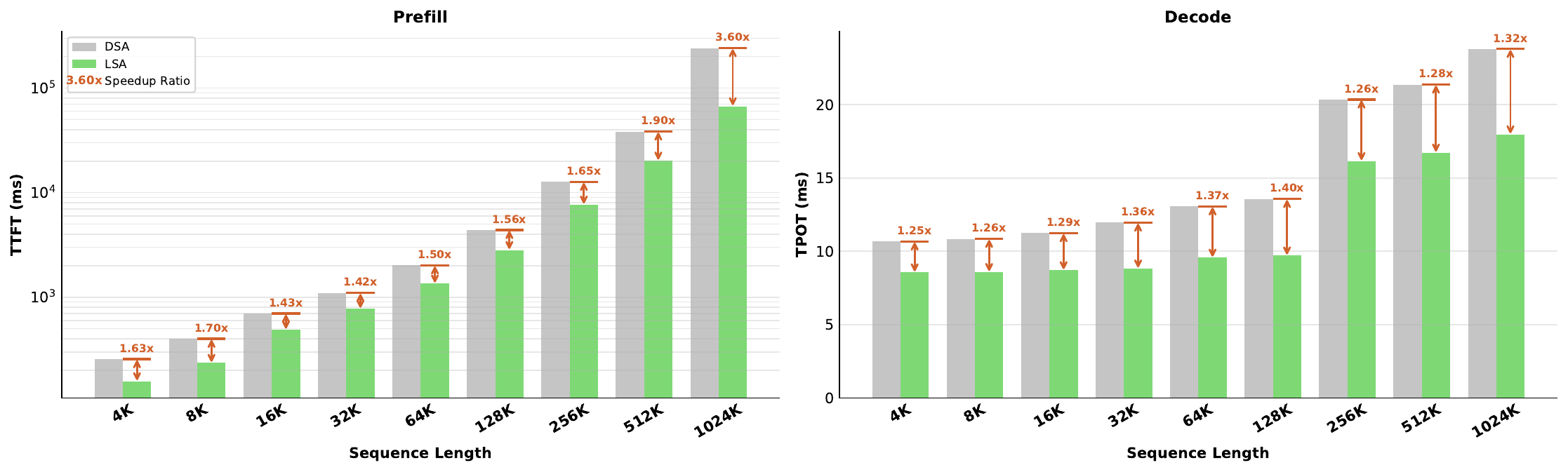}
    \caption{End-to-end inference latency of LSA versus the DSA baseline: prefill time-to-first-token (TTFT, log scale) and decode time-per-output-token (TPOT) across context lengths. The arrow annotates the LSA speedup over DSA.}
    \label{fig:infer_layer}
\end{figure}

As shown in \cref{fig:infer_layer}, LSA achieves a $1.42$--$3.60\times$ prefill speedup and a $1.25$--$1.40\times$ decode speedup over DSA. The prefill speedup increases with context length because the indexer accounts for a growing fraction of latency. Decode speedup peaks at 128K and decreases slightly at 256K and beyond, where the serving configuration switches to KVP. By sharding the KV cache, KVP reduces the indexer's per-rank workload and thus narrows LSA's relative advantage over DSA.

\paragraph{Speedup analysis.}
SFA scales as $\mathcal{O}(L\kmax)$ in prefill and $\mathcal{O}(\kmax)$ in decoding, whereas the indexer scales as $\mathcal{O}(L^2)$ and $\mathcal{O}(L)$, respectively. The indexer therefore accounts for an increasing fraction of latency as the context length grows.
At short context lengths, SFA dominates and most of the gain comes from SI. At long context lengths, the indexer dominates. CLI reduces its cost by reusing indices across layers, while HI mitigates its super-linear prefill scaling beyond 256K.

\paragraph{Compatibility with KV-cache offloading.}
KV-cache offloading complements KVP by storing the cache in host memory and transferring only the chunks accessed by sparse attention. Its efficiency depends on temporal locality across decoding steps. SI improves this locality by reserving half of the attention budget for contiguous sink and sliding-window regions, increasing the average inter-step chunk overlap from $65.05\%$ to $82.04\%$ and reducing per-layer reload latency from $53.88\,\mu$s to $30.46\,\mu$s. CLI further enables asynchronous prefetching for index-reuse layers, reducing the visible latency to $15.23\,\mu$s, or $28\%$ of the DSA baseline.

\paragraph{Integration with multi-step MTP.}
CLI also allows the three MTP steps used for speculative decoding to reuse the indices produced by a single indexer invocation (\cref{sec:mtp_integration}). Since draft tokens are verified by the main model, changes in draft quality affect acceptance length rather than final generation quality. As shown in \cref{tab:mtp_accept}, LSA achieves an average acceptance length of 3.11, compared with 3.15 for dense MLA. The difference indicates negligible impact on speculative-decoding efficiency.

\begin{table}[t]
\centering
\caption{Mean acceptance length of a 3-step MTP module under LSA (DSA with SI and CLI) versus dense MLA, across four representative tasks. Higher is better; the theoretical maximum for 3-step MTP is 4.}
\label{tab:mtp_accept}
\small
\begin{tabular}{l|cccc|c}
\toprule
& HumanEval & GSM8K & AIME & MRCR & \textbf{Avg} \\
\midrule
3-step MTP with LSA & 2.96 & 3.17 & 2.88 & 3.46 & 3.11 \\
3-step MTP with MLA  & 2.97 & 3.20 & 2.83 & 3.59 & 3.15 \\
\bottomrule
\end{tabular}
\end{table}

\section{Experiments and Results}
\label{sec:exp}

\subsection{Setup}
\label{sec:exp_setup}

\paragraph{Model Configuration.}
We validate the effectiveness of LongCat Sparse Attention (LSA) on two model scales from the LongCat family: \longcat (69B-A3B) and LongCat-Flash (560B-A27B). Both adopt the shortcut-connected MoE architecture~\citep{cai2024shortcut} with Multi-head Latent Attention (MLA)~\citep{liu2024deepseekv2}. \longcat contains 14 shortcut layers (each with two serial attention operations, yielding 28 attention layers total) with 32 attention heads; LongCat-Flash consists of 28 shortcut layers (56 attention layers total) with 64 attention heads. We evaluate three attention configurations: (1)~\emph{MLA}: the standard Multi-head Latent Attention baseline with no sparsity; (2)~\emph{DSA}: the standard DeepSeek Sparse Attention with $\kmax = 2048$; and (3)~\emph{LSA}: our proposed method with total $\kmax = 2048$, of which sink $K_{\text{sink}} = 16$ and window $K_{\text{swa}} = 1024$ are fixed, combined with hierarchical indexing and cross-layer indexing ($N = 2$).  For LongCat-Flash, we only compare LSA with MLA due to its larger scale and training resource constraints. Detailed model configurations are listed in \cref{tab:model_config}.

\begin{table}[htbp]
\centering
\caption{Model configurations for the two model scales used in experiments.}
\label{tab:model_config}
\small
\begin{tabular}{lcc}
\toprule
 & \textbf{\longcat} & \textbf{LongCat-Flash} \\
\midrule
\# Parameters (total / active) & 69B / 3B & 560B / 27B \\
\midrule
Layers (shortcut / attention) & 14 / 28 & 28 / 56 \\
Attention heads (core / indexer) & 32 / 16 & 64 / 32 \\
LoRA rank (query / kv) & 1536 / 512 & 1536 / 512 \\
QK head dim (rope / nope) & 64 / 128 & 64 / 128 \\
Indexer head dim (rope / nope) & 64 / 64 & 64 / 64 \\
\midrule
Max Sequence Length      & 512K & 256K \\
\bottomrule
\end{tabular}
\end{table}

\paragraph{Training.}
Starting from mid-training checkpoints, we extend the context window through a two-stage long-context training process. \longcat is trained for 100B tokens at 128K followed by 100B tokens at 512K, while LongCat-Flash is trained for 100B tokens at 128K followed by 20B tokens at 256K. We convert MLA checkpoints to DSA and LSA during the final one-third of long-context training (512K for \longcat and 128K for LongCat-Flash), a transition that begins with a 1,000-step warm-up (7.5B tokens) before entering full sparse training. Ablations (\cref{sec:abl_ctx}) show that the conversion point has no measurable impact on final performance, and the chosen points are primarily for training efficiency. After long-context training, both models undergo supervised fine-tuning with the same settings as their corresponding technical reports~\citep{liu2026scaling,team2025longcat}.

\paragraph{Evaluation Benchmarks.}
We evaluate across two capability groups:
\begin{itemize}
\item \textbf{Long-context}: for chat and thinking models we adopt HELMET~\citep{yen2024helmet}, a comprehensive long-context benchmark covering Recall (RULER~\citep{hsieh2024ruler}), RAG (HotpotQA~\citep{yang2018hotpotqa} + PopQA~\citep{mallen2023popqa}), Re-rank (MS MARCO~\citep{bajaj2016ms}), LongQA (NarrativeQA~\citep{kocisky2018narrativeqa} + $\infty$Bench~\citep{zhang2024infbench}), Citation (ALCE~\citep{gao2023alce}, with ASQA~\citep{stelmakh2022asqa} and QAMPARI~\citep{rubin2022qampari}), and Summarization (Multi-LexSum~\citep{shen2022multilexsum} + $\infty$Bench~\citep{zhang2024infbench}). For base models, which we evaluate throughout our ablation studies (\cref{sec:ablation}), we use the line retrieval task of LongEval~\citep{longchat2023}, an $n$-shot needle-in-a-haystack task that we find effective for distinguishing base model quality.
\item \textbf{Overall}: general capability is assessed with MMLU~\citep{hendrycks2021mmlu}, MMLU-Pro~\citep{wang2024mmlupro}, and CMMLU~\citep{li2023cmmlu}, C-Eval~\citep{huang2024ceval}; reasoning with GPQA-Diamond~\citep{rein2024gpqa}, MATH500~\citep{hendrycks2021math}, and AIME 2024/2025~\citep{aime2024,aime2025}; coding with HumanEval+~\citep{liu2023your}, MBPP+~\citep{liu2023your} and LiveCodeBench~\citep{jain2024livecodebench}.
\end{itemize}

\subsection{Results of LSA}
\label{sec:exp_results}

\paragraph{Long-context evaluation.}
\cref{tab:lsa_longctx} reports HELMET results. On \longcat scale, LSA achieves an average score of 59.02, compared to 58.50 (MLA) and 58.60 (DSA), indicating that LSA retains the long-context capability of full attention despite its sparse indexing. On LongCat-Flash scale, LSA achieves a larger gain over MLA (64.43 vs.\ 62.70), driven primarily by the Re-rank category (+9.3). We attribute this gap to the evaluation setup: LongCat-Flash is a thinking model that tends to produce long reasoning traces. Our analysis reveals that LSA generates slightly shorter outputs than MLA, resulting in a smaller fraction of responses being truncated by the maximum generation length. On the Re-rank subset specifically, MLA produces notably longer generations, leading to more frequent truncation and lower scores. On the remaining categories the two architectures perform comparably.

\paragraph{Overall capability evaluation.}
\cref{tab:lsa_general} compares general knowledge, reasoning, and coding benchmarks. Across both model scales, LSA, DSA, and MLA achieve comparable scores on all evaluated tasks, with no consistent winner among the three. This further confirms that LSA preserves the capability of full attention across different model scales and diverse task types.

\begin{table}[htbp]
\centering
\caption{HELMET evaluation of LSA vs.\ MLA on \longcat (69B-A3B) and LongCat-Flash (560B-A27B). Each category score is the average across its constituent benchmarks.}
\label{tab:lsa_longctx}
\small
\begin{tabular}{ll|cccccc|@{\hskip 10pt}c}
\toprule
\textbf{Model} & \textbf{Attn} & Recall & RAG & Re-rank & LongQA & Cite & Summ & \textbf{Avg} \\
\midrule
\multirow{3}{*}{\shortstack[l]{\longcat\\(69B-A3B, Chat)}} & MLA & 98.83 & 64.61 & 71.34 & 44.03 & 35.83 & 36.38 & 58.50 \\
 & DSA & \textbf{99.13} & \textbf{65.10} & 70.73 & 42.98 & 36.91 & \textbf{36.78} & 58.60 \\
 & LSA & 98.63 & 64.38 & \textbf{72.64} & \textbf{44.46} & \textbf{37.53} & 36.48 & \textbf{59.02} \\
\midrule
\multirow{2}{*}{\shortstack[l]{LongCat-Flash\\(560B-A27B, Thinking)}} & MLA & 97.30 & \textbf{85.40} & 62.09 & 38.89 & 43.98 & 48.53 & 62.70 \\
 & LSA & \textbf{97.38} & 84.60 & \textbf{71.36} & \textbf{38.97} & \textbf{45.20} & \textbf{49.10} & \textbf{64.43} \\

\bottomrule
\end{tabular}
\end{table}

\begin{table}[htbp]
\centering
\caption{Standard benchmark evaluation of LSA vs.\ MLA and DSA on \longcat (69B-A3B) and LongCat-Flash (560B-A27B).}
\label{tab:lsa_general}
\small
\setlength{\tabcolsep}{5pt}
\begin{tabular}{ll ccc @{} p{16pt} cc}
\toprule
 & & \multicolumn{3}{c}{\textbf{\longcat (69B-A3B)}} & & \multicolumn{2}{c}{\textbf{LongCat-Flash (560B-A27B)}} \\
\cmidrule(lr){3-5} \cmidrule(lr){7-8}
\textbf{Category} & \textbf{Benchmark} & \textbf{MLA} & \textbf{DSA} & \textbf{LSA} & & \textbf{MLA} & \textbf{LSA} \\
\midrule
\multirow{4}{*}{General}
 & MMLU      & \textbf{85.54} & 85.27 & 85.50 & & 89.87 & \textbf{90.05} \\
 & MMLU-Pro  & 78.39 & \textbf{79.68} & 78.33 & & \textbf{82.65} & 82.39 \\
 & CMMLU     & 82.46 & 81.82 & \textbf{82.61} & & 87.02 & \textbf{87.23} \\
 & C-Eval    & \textbf{86.33} & 86.31 & 85.81 & & 88.38 & \textbf{89.50} \\
\midrule
\multirow{4}{*}{Reasoning}
 & GPQA-Diamond & 68.72 & 68.66 & \textbf{69.51} & & \textbf{83.49} & 83.46 \\
 & MATH500     & \textbf{98.20} & 96.60 & 97.20 & & 98.00 & \textbf{98.40} \\
 & AIME 2024   & 72.60 & 71.87 & \textbf{73.44} & & 90.83 & \textbf{91.04} \\
 & AIME 2025   & 59.90 & 63.65 & \textbf{64.27} & & 89.89 & \textbf{90.00} \\
\midrule
\multirow{3}{*}{Coding}
 & HumanEval+      & \textbf{86.59} & 83.54 & \textbf{86.59} & & 70.73 & \textbf{71.95} \\
 & MBPP+           & 78.31 & 79.10 & \textbf{79.37} & & \textbf{72.75} & 71.69 \\
 & LiveCodeBench   & 41.19 & \textbf{42.07} & 41.85 & & \textbf{80.40} & 79.74 \\
\bottomrule
\end{tabular}
\end{table}

\subsection{Ablation Studies}
\label{sec:ablation}

All ablations are performed at the \longcat (69B-A3B) scale. LSA variants are compared with the MLA baseline through training/validation loss and Needle-in-a-Haystack evaluation, followed by HELMET evaluation after supervised fine-tuning.

\subsubsection{Streaming-Aware Indexing Preserves Quality with Half Fixed Budget}
\label{sec:abl_streaming}

\begin{figure}[htbp]
    \centering
    \includegraphics[width=\linewidth]{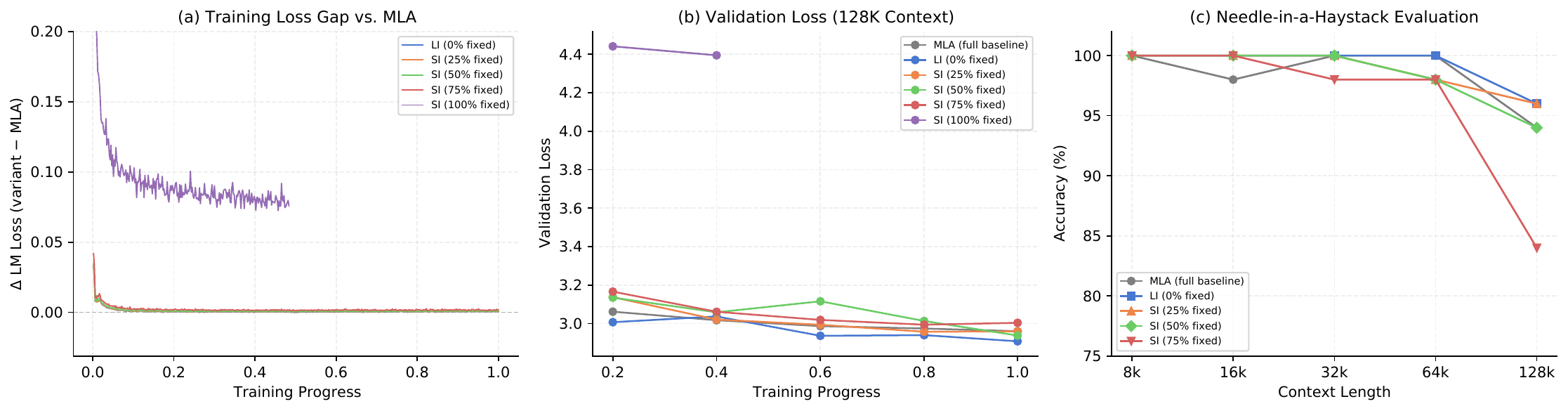}
    \caption{Streaming-Aware Indexing ablation. ``X\% fixed'' = fraction of budget $K$ allocated to the fixed window and sink tokens. (a)~Training loss gap relative to MLA. (b)~Validation loss on long-context set. (c)~Needle-in-a-haystack retrieval accuracy.}
    \label{fig:streaming_ablation}
\end{figure}

\begin{table}[htbp]
\centering
\small
\caption{HELMET evaluation for Streaming-Aware Indexing ablation (chat models).}
\label{tab:streaming_chat_eval}
\begin{tabular}{l|cccccc|@{\hskip 10pt}c}
\toprule
 & Recall & RAG & Re-rank & LongQA & Cite & Summ & \textbf{Avg} \\
\midrule
MLA                    & 95.68 & 64.23 & 67.33 & 39.63 & 35.13 & 33.30 & 55.88 \\
LI (0\% fixed)         & \textbf{96.28} & \textbf{66.28} & 65.61 & \textbf{39.98} & 34.66 & 33.78 & 56.10 \\
SI (50\% fixed)        & 94.20 & 65.68 & \textbf{68.08} & 39.33 & \textbf{37.35} & \textbf{34.91} & \textbf{56.59} \\
\bottomrule
\end{tabular}
\end{table}

A larger fixed window improves hardware efficiency through contiguous memory access, but reduces the tokens freely chosen by the indexer, risking degraded long-context modeling. We ablate this trade-off by varying the fraction of the total token budget $K$ allocated to the fixed window (denoted ``X\% fixed'', including 16 sink tokens). Since long-context training is computationally expensive, we train MLA and DSA variants with different fixed ratios (0\%, 25\%, 50\%, 75\%, 100\%) at 128K sequence length and compare their training dynamics and downstream quality. For brevity, we denote DSA with the standard Lightning Indexer as \textbf{LI}, and DSA with the Streaming-Aware Indexer as \textbf{SI}.

\cref{fig:streaming_ablation} summarizes the results. Panel~(a) plots each variant's training loss gap relative to MLA: configurations at 75\% fixed or below show negligible difference throughout training, whereas 100\% fixed (i.e., pure window attention) exhibits a significantly higher loss. Panel~(b) shows validation loss on a long-context set, where 100\% fixed again shows significantly higher loss and was therefore abandoned. Panel~(c) reports needle-in-a-haystack evaluation (LongEval line retrieval task, which we find discriminative for base models) on final checkpoints: at 128K context, 75\% fixed suffers a clear accuracy drop while 0\%--50\% fixed all match or slightly exceed MLA. We therefore discard the 75\% setting and apply supervised fine-tuning to the remaining groups. \cref{tab:streaming_chat_eval} evaluates the resulting chat models on HELMET: SI (50\% fixed) achieves comparable performance to both LI (0\% fixed) and MLA. We select 50\% fixed (a $\sim$1:1 fixed-to-dynamic ratio) as the default, as it maximizes the fixed window size without degrading long-context quality.

\subsubsection{Cross-Layer Sharing Preserves Quality Up to \texorpdfstring{$N{=}2$}{N=2}}
\label{sec:abl_cli}

\begin{figure}[htbp]
    \centering
    \includegraphics[width=\linewidth]{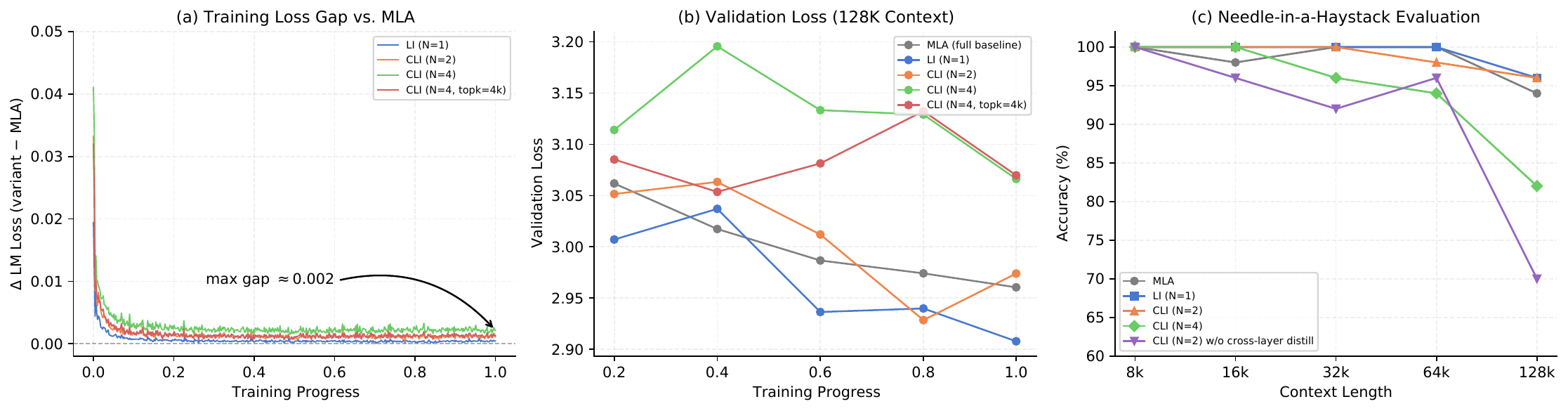}
    \caption{Cross-Layer Indexing ablation (base models). (a)~Training loss gap relative to MLA. (b)~Long-context validation loss. (c)~Needle-in-a-haystack retrieval accuracy across context lengths.}
    \label{fig:cli_ablation_combined}
\end{figure}

\begin{table}[htbp]
\centering
\small
\caption{HELMET chat evaluation for Cross-Layer Indexing ablation. $N=2$ performs on par with MLA and the LI baseline ($N=1$), confirming that halving the indexer passes preserves quality.}
\label{tab:cli_chat_eval}
\begin{tabular}{l|cccccc|@{\hskip 10pt}c}
\toprule
 & Recall & RAG & Re-rank & LongQA & Cite & Summ & \textbf{Avg} \\
\midrule
MLA             & 95.68 & 64.23 & 67.33 & 39.63 & \textbf{35.13} & 33.30 & 55.88 \\
LI ($N{=}1$)   & \textbf{96.28} & \textbf{66.28} & 65.61 & 39.98 & 34.66 & 33.78 & \textbf{56.10} \\
CLI ($N{=}2$)  & 93.05 & 65.80 & \textbf{68.13} & \textbf{40.45} & 33.30 & \textbf{33.92} & 55.78 \\
\bottomrule
\end{tabular}
\end{table}

We study the impact of CLI group size \(N\) under uniform layer partitioning. Increasing \(N\) reduces indexer computation by allowing more layers to share one indexer, but also makes the shared indexer responsible for a broader range of layer-wise saliency patterns. Following \cref{sec:abl_streaming}, we train all variants at 128K sequence length and compare MLA against DSA with CLI group sizes \(N\in\{1,2,4\}\), where \(N{=}1\) corresponds to the standard Lightning Indexer (LI). To examine whether a larger selection budget can compensate for more aggressive sharing, we additionally evaluate \(N{=}4\) with the top-\(k\) budget increased from 2K to 4K. We denote DSA with the standard indexer as \textbf{LI} and DSA with CLI as \textbf{CLI}; unless specified otherwise, all variants use the default top-\(k\) budget of 2K.

\cref{fig:cli_ablation_combined} summarizes the training and evaluation results. Panel~(a) shows the training loss gap relative to MLA. Although CLI (\(N{=}4\)) exhibits a slightly larger gap, the final difference remains below 0.002 in absolute loss (less than 0.5\% relative to the overall loss), suggesting that training loss alone is insufficient to capture long-context quality. This is likely because the variable-length training mixture contains many short sequences where sparse indexing differences have limited impact.

Panel~(b) evaluates validation loss on a long-context set. While the curves show some variance, \(N{=}4\) consistently underperforms at the end of training, and increasing the top-\(k\) budget to 4K fails to recover the degradation. This indicates that the challenge of sharing one indexer across four layers cannot be addressed simply by expanding the selection budget.

Panel~(c) evaluates the final base-model checkpoints using needle-in-a-haystack retrieval (omitting the \(N{=}4\), topk=4k variant since its validation behavior is similar to \(N{=}4\)). Beyond 32K context, \(N{=}4\) suffers a clear accuracy degradation that grows with sequence length, whereas \(N{=}1\) and \(N{=}2\) remain comparable to or slightly better than MLA up to 128K.

Based on these results, we discard \(N{=}4\) and perform supervised fine-tuning on the remaining configurations. \cref{tab:cli_chat_eval} shows that CLI (\(N{=}2\)) achieves comparable HELMET performance to both LI (\(N{=}1\)) and MLA, demonstrating that halving the number of indexer computations preserves long-context quality. We therefore adopt \(N{=}2\) as the default CLI group size.

\subsubsection{Cross-Layer Distillation Is Essential for Index Reuse}
\label{sec:abl_cli_distill}

As shown in \cref{fig:cli_overlap}, adjacent layers exhibit substantial overlap in their salient token sets. This raises a key question: is such inherent inter-layer consistency sufficient for cross-layer index reuse, or is the cross-layer distillation loss (\cref{sec:cli}), which explicitly trains the owner indexer to serve multiple layers within a group, necessary?

This ablation requires no additional training. Since the indexer input is detached from the model computation graph during training (\cref{sec:dsa_training}), indexers at different layers are optimized independently. We therefore construct a ``CLI (N=2) w/o cross-layer distill'' variant by removing redundant indexers from a trained LI (N=1) model and sharing the remaining indexer across two layers at inference time.

We evaluate this variant on needle-in-a-haystack retrieval. As shown in \cref{fig:cli_ablation_combined}(c), removing cross-layer distillation causes severe degradation at long contexts: accuracy drops to 70\% at 128K, below both distilled CLI (N=4, 82\%) and distilled CLI (N=2, 96\%). These results show that naive index reuse is insufficient. Cross-layer distillation is essential because it explicitly trains the owner indexer to capture the joint saliency patterns of all layers in the group, rather than producing indices optimized for only a single layer.

\subsubsection{CLI Extends to MTP Layers}
\label{sec:abl_cli_mtp}

\begin{figure}[htbp]
    \centering
    \includegraphics[width=0.95\linewidth]{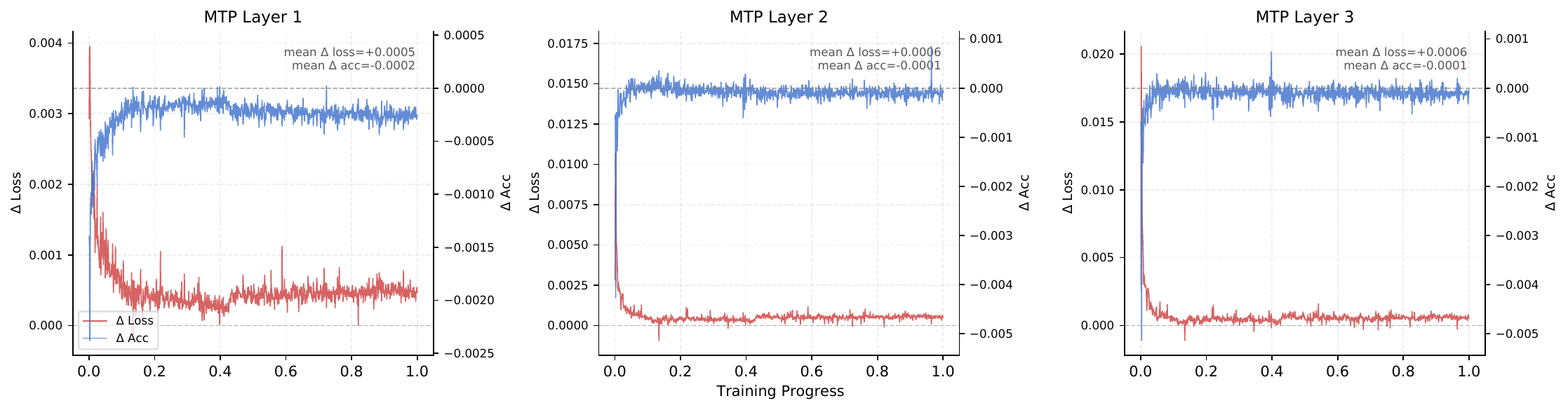}
    \caption{MTP metrics delta (DSA w/ CLI $-$ MLA) for all 3 MTP steps. Accuracy and loss differences remain tightly centered around zero throughout training, confirming that CLI across MTP steps preserves prediction quality.}
    \label{fig:cli_mtp_ablation}
\end{figure}

We validate our design in \cref{sec:cli} by comparing MLA with 3-step MTP against DSA with CLI shared across all 3 MTP steps (\(N{=}3\)). Since MTP outputs are draft proposals verified by the main model, any degradation appears as reduced acceptance length. As shown in \cref{fig:cli_mtp_ablation}, after sparse training converges, the LM loss gap remains below \(10^{-3}\) and prediction accuracy differences stay within \(0.1\%\) across all MTP steps. Consistently, \cref{tab:mtp_accept} shows nearly identical acceptance lengths between LSA and MLA across HumanEval, GSM8K, AIME, and MRCR (\(3.11\) vs. \(3.15\)). These results demonstrate that CLI reuse across MTP steps preserves draft quality while eliminating redundant indexer computation.

\subsubsection{Hierarchical Indexing Preserves Quality with Proper Configuration}
\label{sec:abl_hier}
Hierarchical Indexing is training-free (\cref{sec:hierarchical}): it is applied purely at inference time and introduces no change to the trained model. Its only quality risk is insufficient coarse-stage recall, which the fine stage cannot recover. In this section we ablate the three knobs that govern this approximation---the pooling method, which indexer layers it is applied to, and the coarse-level recall budget---to identify the configuration that minimizes quality loss while retaining its speedup.
\paragraph{Pooling method.}
The previous work AsyncTLS~\citep{hu2026asynctls} applied max-pooling to compute the block-level score, and the concurrent work HISA~\citep{xu2026hisa} also suggests max-pooling as a future direction; we compared the mean pooling to the max-pooling option. Specifically, instead of the mean-pooled dot product $\mathbf{q}_{t,j}^I \cdot \mathbf{k}_{n}^{\mean}$ in \cref{eq:block_score}, for each sub-block $n$ we precompute the per-dimension extrema of its constituent keys, $k_{n,d}^{\max} = \max_{s \in \text{sub-block}_n} k_{s,d}^I$ and $k_{n,d}^{\min} = \min_{s \in \text{sub-block}_n} k_{s,d}^I$, and replace the ReLU argument with a block-level score that upper-bounds the maximum token score within that sub-block:
\begin{equation}
\label{eq:sub-block_score_minmax}
\tilde{s}_{t,n,j} = \sum_{d=1}^{D}\max\!\left(q_{t,j,d}^I \cdot k_{n,d}^{\max},\; q_{t,j,d}^I \cdot k_{n,d}^{\min}\right),
\end{equation}
where $D$ is the indexer head dimension, so the coarse score becomes $I_{t,p}^{\text{page}} = \sum_{j=1}^{H^I} w_{t,j}^I \sum_{n=1}^{N} \mathrm{ReLU}(\tilde{s}_{t,n,j})$.
We conduct ablation studies for the pooling method across different pooling sizes under the NIAH 128K task. Mean pooling with a pooling size of 8 is the optimal choice, considering the effectiveness-efficiency tradeoff, as shown in \cref{tab:HI_ablations_pooling}.

\begin{table}[!h]
\centering
\small
\caption{Ablation studies for the different pooling methods and the pooling sizes under the Needle-in-a-haystack (NIAH) 128K task, the coarse level recalls $M=128$ pages (16K tokens). The selected configuration is marked in bold.}
\label{tab:HI_ablations_pooling}
\begin{tabular}{l|ccccc}
\toprule
& \multicolumn{5}{c}{\textbf{NIAH 128K}} \\
\midrule
Method / Size & 1 & 4 & \textbf{8} & 16 & 32 \\
\midrule
\textbf{Mean} & 82 & 76 & \textbf{80} & 78 & 74 \\
MinMax &82 & 64 & 60 & 68 & 68 \\
\bottomrule
\end{tabular}
\end{table}

\paragraph{Layer selection.}
We conduct ablation studies to choose which indexer to turn off the HI. The ablation shows that the shallow layers are more sensitive to the possible recalling error introduced by the HI. The final choice is to turn off the first 4 indexers to help preserve the model performance, as shown in \cref{tab:pooling_layer_ablations}.

\begin{table}[h]
\centering
\small
\caption{Ablation studies for turning off the HI of which indexers under the Needle-in-a-haystack (NIAH) 128K task. Mean Pooling with a pooling blocksize $B$ of 8, recalling $M=256$ pages (32K tokens). The selected configuration is marked in bold.}
\label{tab:pooling_layer_ablations}
\begin{tabular}{l|c}
\toprule
\textbf{Options} & \textbf{NIAH 128K} \\
\midrule
No turning off HI & 84  \\
Turn off HI for the first 2 indexers & 84 \\
Turn off HI for the first 3 indexers & 86 \\
\textbf{Turn off HI for the first 4 indexers} & \textbf{92} \\
Turn off HI for the first 2 indexers and the last 2 indexers & 84 \\
\bottomrule
\end{tabular}
\end{table}

\paragraph{Recall budget.}
To determine the recall pages of the coarse-level indexer of HI, we compared the performance of the different recall numbers in \cref{tab:pooling_topk_comparison}. Recalling $M=1024$ pages is the boundary choice to preserve the performance in the MRCR tasks.
\begin{table}[h]
\centering
\small
\caption{Ablation studies for the different number of recalled pages under MRCR tasks. Mean Pooling with a pooling blocksize of 8, turning off HI for the first 4 indexers. The selected configuration is marked in bold.}
\label{tab:pooling_topk_comparison}
\begin{tabular}{l|cc}
\toprule
\textbf{TopK of Pages} & \textbf{MRCR 256K} & \textbf{MRCR 512K} \\
\midrule
256 (32K)  & 30.96 & 24.32 \\
512 (64K)  & 30.17 & 23.09 \\
\textbf{1024 (128K)}  & \textbf{32.34} & \textbf{30.28} \\
\midrule
LSA baseline without HI & 31.49 & 27.07 \\
\bottomrule
\end{tabular}
\end{table}

\paragraph{Optimal setting.}
Based on the results of the three ablation studies, to preserve the lossless performance, the final optimal setting is the Mean pooling method with a pooling blocksize of 8, and turning off HI for the first 4 indexers and recalling $M=1024$ pages ($128K$ tokens) for all long context tasks with sequence lengths $\ge256$K. At the operator level, HI under this optimal setting achieves up to a $4.11\times$ indexer speedup over the baseline at 1024K (\cref{tab:hi_infer}). We also conduct a comprehensive evaluation for the Hierarchical Indexer with this optimal setting, across various long context tasks with sequence lengths up to 1024K in \cref{tab:release_longctx}.

\subsubsection{LSA Is Robust to Conversion Timing}
\label{sec:abl_ctx}

\begin{figure}[htbp]
    \centering
    \includegraphics[width=\linewidth]{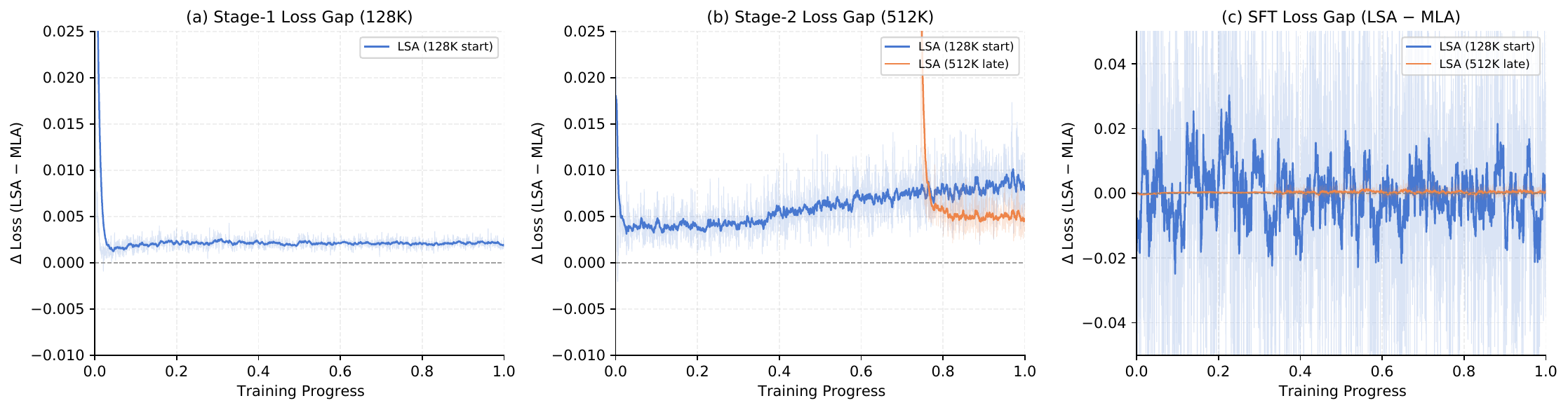}
    \caption{Context-length extension ablation. (a)~128K-stage loss gap (LSA $-$ MLA). (b)~512K-stage loss gap (LSA $-$ MLA). (c)~SFT loss gap (LSA $-$ MLA). Values near zero indicate negligible quality difference.}
    \label{fig:ctx_train_ablation}
\end{figure}

\begin{table}[htbp]
\centering
\small
\caption{HELMET evaluation for context-extension strategies. Both early (128K start) and late (512K start) LSA introduction match the MLA baseline across all long-context categories.}
\label{tab:ctx_chat_eval}
\begin{tabular}{l|cccccc|@{\hskip 10pt}c}
\toprule
 & Recall & RAG & Re-rank & LongQA & Cite & Summ & \textbf{Avg} \\
\midrule
MLA (baseline)         & 98.83 & \textbf{64.61} & 71.34 & 44.03 & 35.83 & 36.38 & 58.50 \\
LSA (128K start) & \textbf{99.35} & 64.07 & 71.77 & 44.05 & 37.51 & \textbf{37.01} & 58.96 \\
LSA (512K late)  & 98.63 & 64.38 & \textbf{72.64} & \textbf{44.46} & \textbf{37.53} & 36.48 & \textbf{59.02} \\
\bottomrule
\end{tabular}
\end{table}
Following standard practice, LSA is converted from a well-trained MLA model during long-context extension. Since LSA outpaces MLA beyond ${\sim}128$K (\cref{app:train_vs_mla}), throughput-optimal training dictates converting at 128K and remaining sparse thereafter. However, prior works~\citep{deepseek2025dsa,zeng2026glm5} defer conversion to the \emph{final} extension stage. We hypothesize this stems from indexer distribution shifts: in sparse training, the KL loss supervision couples to the indexer's own top-$\kmax$ token selection ($\mathcal{S}_t$, \cref{eq:sparse_loss}). Abrupt context-length transitions force the indexer through out-of-distribution shifts, potentially degrading supervision quality if converted too early.

To verify whether the conversion timing affects quality, we compare two schedules on \longcat: (1)~\textbf{LSA (128K start)}, which converts from MLA to LSA at the very beginning of the 128K stage (100B tokens) and then completes the entire 512K stage (100B tokens) in sparse mode; and (2)~\textbf{LSA (512K late)}, which keeps full MLA throughout the 128K stage and converts to LSA only in the final third of the 512K stage---following the standard late-conversion recipe above.

As shown in \cref{fig:ctx_train_ablation}, while the late-conversion schedule maintains a slightly tighter pre-training loss gap relative to MLA, both schedules keep this gap strictly below $0.01$ throughout ($<0.5\%$ of the absolute loss, ${\sim}2$), and the SFT loss gap is centered at zero. Downstream evaluation on HELMET (\cref{tab:ctx_chat_eval}) confirms that early conversion achieves performance parity with both late conversion and full MLA (58.96 vs.\ 59.02 and 58.50, respectively). Since downstream performance is closely matched, we recommend converting to LSA as early as the 128K crossover stage to maximize training efficiency.

\section{LongCat-Flash-Lite-Sparse}
\label{sec:release_model}
Building on the techniques developed in this work, we introduce \textbf{LongCat-Flash-Lite-Sparse}, which integrates the full LSA recipe into \longcat~\citep{liu2026scaling}. Compared with the previously released model with full attention, the new model replaces dense MLA with LSA for substantially improved inference efficiency, extends the native context length from 128K to 1M, achieves stronger agentic capabilities, and preserves strong reasoning and general-knowledge performance.

\subsection{Model Informations}
\label{sec:release_info}

\paragraph{Architecture.}
The released model is based on \longcat (69B-A3B) and replaces dense MLA attention with LSA using the configuration validated in \cref{sec:exp_setup}: a total attention budget of \(\kmax=2048\) tokens, including \(K_{\text{sink}}=16\) sink tokens, \(K_{\text{swa}}=1024\) sliding-window tokens, and dynamically selected sparse tokens. The model uses Cross-Layer Indexing with group size \(N=2\), while Hierarchical Indexing is applied as a training-free inference optimization beyond its efficiency crossover point. We further integrate a 3-step MTP module for speculative decoding, where all MTP steps share one indexer through CLI (\(N=3\)).

\paragraph{Training pipeline.}
We extend \longcat's long-context training pipeline through five stages: 32K, 64K, 128K, 256K, and 1M. Based on profiling and ablation studies, we convert dense MLA to LSA at the beginning of the 128K stage, followed by sparse training across the 128K, 256K, and 1M stages. The MTP module is introduced at the 32K stage and converted to LSA together with the main model at 128K. After long-context extension, the model undergoes supervised fine-tuning. Note that only Streaming-Aware Indexing and Cross-Layer Indexing participate in training; Hierarchical Indexing is training-free and applied only during inference.

\subsection{Model Evaluation}
\label{sec:release_eval}
We first evaluate long-context capabilities with and without HI (\cref{tab:release_longctx}), followed by a comprehensive evaluation on agentic, general-knowledge, and reasoning tasks (\cref{tab:release_model_eval}). For clarity, in these comparisons, \textbf{Lite-Dense} denotes \longcat, while \textbf{Lite-Sparse (w/o HI)} and \textbf{Lite-Sparse (w/ HI)} denote the sparse variants evaluated without and with Hierarchical Indexing, respectively.

\paragraph{Benchmarks.}
We evaluate across four capability groups:
\begin{itemize}
\item \textbf{Long-context}: We adopt ATLAS~\citep{huang2026atlas}, which unifies existing long-context benchmarks under a common length axis and evaluates them at 8K, 16K, 32K, 64K, 128K, 256K, 512K, and 1M tokens. The evaluation covers retrieval (MRCR 8-needle~\citep{openai2025mrcr}), aggregation (OOLong-Synth~\citep{bertsch2025oolong}), multi-step reasoning (GraphWalks~\citep{graphwalks}), question answering (LOFT-Text Retrieval~\citep{loft}), in-context learning (HELMET-ICL~\citep{yen2024helmet}), code understanding (LongCodeQA~\citep{rando2025longcodebench}), long-range memory (AMemBench-ACU~\citep{jiayang2026amemgym}), and holistic evaluation (LongBench-v2~\citep{bai2024longbenchv2} and AA-LCR~\citep{artificialanalysis2025lcr}).
\item \textbf{Agentic}: We evaluate agentic coding (SWE-Bench Verified, SWE-Bench Pro, SWE-Bench Multilingual~\citep{jimenez2023swebench}, and TerminalBench~\citep{merrill2026terminalbench}), agentic tool use ($\tau^2$-Bench~\citep{tau2bench}, VitaBench~\citep{he2025vitabench}, and MCP-Atlas~\citep{bandi2026mcpatlas}), and agentic search (BrowseComp~\citep{wei2025browsecomp,zhou2025browsecompzh} and RWSearch~\citep{rwsearch2026}).
\item \textbf{General Domains}: MMLU~\citep{hendrycks2021mmlu}, MMLU-Pro~\citep{wang2024mmlupro}, CMMLU~\citep{li2023cmmlu}, and C-Eval~\citep{huang2024ceval}.
\item \textbf{Mathematical Reasoning}: GPQA-Diamond~\citep{rein2024gpqa}, MATH500~\citep{hendrycks2021math}, AIME 2026~\citep{dekoninck2026matharena}, HMMT 2026~\citep{dekoninck2026matharena}, BeyondAIME~\citep{bytedance2025beyondaime}, and IMO AnswerBench~\citep{luong2025imoanswerbench}.
\end{itemize}

\paragraph{Long-context evaluation.}

\cref{tab:release_longctx} compares Lite-Sparse with and without HI on nine long-context benchmarks up to 1M context lengths. Overall, HI maintains highly competitive performance across all tasks. On most benchmarks, the score variations remain within roughly 1 point (e.g., LongBench-v2: \(53.64\) vs.\ \(52.50\), MRCR: \(44.47\) vs.\ \(44.66\), and AMemBench-ACU: \(33.13\) vs.\ \(33.25\)), with a slight drop observed on code understanding (LongCodeQA: \(59.37\) vs.\ \(62.30\)). These results demonstrate a favorable efficiency-accuracy trade-off: training-free Hierarchical Indexing largely preserves long-context capabilities while delivering substantial inference speedups with only minor quality degradation.

\begin{table}[htbp]
\centering
\caption{Long-context evaluation of LongCat-Flash-Lite-Sparse on the ATLAS~\citep{huang2026atlas} benchmark suite, organized by capability dimension. Every component supports context lengths up to 1M tokens.}
\label{tab:release_longctx}
\small
\setlength{\tabcolsep}{10pt}
\begin{tabular}{lll|cc}
\toprule
\textbf{Layer} & \textbf{Dimension} & \textbf{Benchmark} & \textbf{Lite-Sparse (w/o HI)} & \textbf{Lite-Sparse (w/ HI)} \\
\midrule
\multirow{3}{*}{Foundational}
 & Retrieval     & MRCR (8-needle)                     & \textbf{44.66} & 44.47 \\
 & Aggregation   & OOLong-Synth                        & \textbf{38.42} & 37.88 \\
 & Multi-step Reasoning & GraphWalks Extend            & \textbf{66.27} & 65.63 \\
\midrule
\multirow{6}{*}{Application}
 & Question Answering & LOFT Retrieval Extend          & 43.75 & \textbf{44.38} \\
 & In-context Learning & HELMET-ICL Extend            & \textbf{91.63} & 90.50 \\
 & Code Understanding & LongCodeQA                     & \textbf{62.30} & 59.37 \\
 & Long-range Memory & AMemBench-ACU                   & \textbf{33.25} & 33.13 \\
 & \multirow{2}{*}{Holistic Assessment}
                 & LongBench-v2                        & 52.50 & \textbf{53.64} \\
 &               & AA-LCR                              & \textbf{48.00} & 47.33 \\
\bottomrule
\end{tabular}
\end{table}

\begin{table}[htbp]
\centering
\caption{Evaluation results of LongCat-Flash-Lite-Sparse compared with \longcat.}
\label{tab:release_model_eval}
\small
\setlength{\tabcolsep}{8pt}
\begin{tabular}{l|c|cc}
\toprule
\textbf{Benchmark} & \textbf{Lite-Dense} & \textbf{Lite-Sparse (w/o HI)} & \textbf{Lite-Sparse (w/ HI)} \\
\midrule
\multicolumn{4}{c}{\textbf{Agentic Coding}} \\
\midrule
SWE-Bench Verified $_{\text{(acc)}}$          & 54.40 & \textbf{68.20} & 65.20 \\
SWE-Bench Pro $_{\text{(acc)}}$               & -- & \textbf{40.63} & 39.40 \\
SWE-Bench Multilingual $_{\text{(acc)}}$      & 38.10 & \textbf{59.33} & 56.00 \\
TerminalBench 2.0 $_{\text{(acc)}}$           & \textbf{33.75} & 33.70 & 32.58 \\
\midrule
\multicolumn{4}{c}{\textbf{Agentic Tool Use}} \\
\midrule
$\tau^2$-Telecom $_{\text{(avg@4)}}$  & 72.80 & 95.18 & \textbf{96.05} \\
VitaBench $_{\text{(avg@4)}}$                 & 7.00 & \textbf{21.67} & 20.42 \\
MCP-Atlas                                      & -- & \textbf{45.60} & 45.00 \\
\midrule
\multicolumn{4}{c}{\textbf{Agentic Search}} \\
\midrule
BrowseComp $_{\text{(pass@1)}}$               & -- & \textbf{48.62} & 48.18 \\
BrowseComp-zh $_{\text{(pass@1)}}$            & -- & \textbf{61.94} & 61.59 \\
RWSearch $_{\text{(pass@1)}}$                 & -- & \textbf{68.50} & 66.00 \\
\midrule
\multicolumn{4}{c}{\textbf{General Domains}} \\
\midrule
MMLU $_{\text{(acc)}}$           & \textbf{85.52} & 85.31 & 85.14 \\
MMLU-Pro $_{\text{(acc)}}$      & 78.29 & \textbf{79.24} & 78.68 \\
CMMLU $_{\text{(acc)}}$         & 82.48 & 84.25 & \textbf{84.51} \\
C-Eval $_{\text{(acc)}}$        & \textbf{86.55} & 85.76 & 85.71 \\
\midrule
\multicolumn{4}{c}{\textbf{Mathematical Reasoning}} \\
\midrule
GPQA-Diamond $_{\text{(avg@16)}}$ & 66.78 & \textbf{69.49} & 69.03 \\
MATH500 $_{\text{(acc)}}$       & \textbf{96.80} & 95.80 & \textbf{96.80} \\
AIME 2026 $_{\text{(avg@32)}}$  & -- & \textbf{65.73} & 64.90 \\
HMMT 2026 Feb $_{\text{(avg@32)}}$ & -- & 40.53 & \textbf{41.47} \\
BeyondAIME $_{\text{(avg@10)}}$ & -- & \textbf{44.20} & 42.30 \\
IMO AnswerBench $_{\text{(avg@4)}}$ & -- & \textbf{49.38} & 46.69 \\
\bottomrule
\end{tabular}
\end{table}

\paragraph{Overall capability evaluation.}
\cref{tab:release_model_eval} summarizes the overall capability of Lite-Sparse compared with Lite-Dense, whose scores are taken from its technical report~\citep{liu2026scaling}.
Lite-Sparse achieves strong performance on agentic benchmarks, including \(68.20\) on SWE-Bench Verified, \(59.33\) on SWE-Bench Multilingual, \(95.18\) on \(\tau^2\)-Telecom, and \(21.67\) on VitaBench (without HI). Enabling HI introduces a modest quality trade-off on several agentic tasks (e.g., SWE-Bench Verified: \(68.20\rightarrow65.20\), SWE-Bench Multilingual: \(59.33\rightarrow56.00\), and RWSearch: \(68.50\rightarrow66.00\)). Overall, Lite-Sparse preserves the dense model's reasoning and general-knowledge capabilities while improving its agentic performance and enabling substantially more efficient long-context inference.

\section{Conclusion, Limitations, and Future Directions}
We presented LongCat Sparse Attention (LSA), a co-designed sparse-attention framework that addresses the inefficient memory access and high indexing overhead of DSA through three complementary mechanisms: \emph{Streaming-Aware Indexing}, \emph{Cross-Layer Indexing}, and \emph{Hierarchical Indexing}. Evaluations at both model scales show that LSA achieves nearly lossless performance relative to full attention while delivering substantial training and inference speedups. The resulting training-efficiency gains make native training with context lengths of up to one million tokens practical and support the development of LongCat-2.0 (1.6T-A48B). To facilitate further research, we also release LongCat-Flash-Lite-Sparse (69B-A3B), an open-source model that integrates LSA into LongCat-Flash-Lite and incorporates an updated long-context training corpus.

A primary constraint of LSA is that while it substantially reduces attention computation, it leaves the total KV-cache footprint intact, as every token must still store a KV entry. Although KV-cache partitioning and host-memory offloading (\cref{sec:inference_speedup}) mitigate per-device memory pressure, they do not lower the aggregate storage overhead. A promising future direction is to combine LSA with complementary KV-cache compression paradigms. For instance, Cross-Layer Attention (CLA)~\citep{brandon2024cla} shares KV states across layers to compress the cache along the depth dimension, whereas DeepSeek-V4’s Compressed Sparse Attention (CSA)~\citep{xu2026deepseek} achieves sequence-dimension compression via block-level sparse selection. Fusing LSA with these orthogonal techniques holds great potential for scaling long-context models that are simultaneously compute- and memory-efficient.

\clearpage
\section{Acknowledgement}
We extend our sincere gratitude to both the infrastructure team and evaluation team for their invaluable support and constructive feedback throughout this project. The primary contributors from these teams include:

\begin{tabular}{p{0.25\textwidth}p{0.25\textwidth}p{0.25\textwidth}p{0.25\textwidth}}
Yuxuan Hu & Gang Liu & Li Wei & Hongjun Wu \\
Jiaxin Hou & Yuwei Jiang & Bole Zhou & Pingwei Sun  \\
Yuanshuo Wang & Xing Hu & Rumei Li & Dengchang Zhao
\end{tabular}

\clearpage

\bibliographystyle{unsrtnat}
\bibliography{references}

\clearpage

\appendix

\section{Training Efficiency of LSA versus Dense MLA}
\label[appendix]{app:train_vs_mla}

\begin{figure}[H]
    \centering
    \includegraphics[width=\linewidth]{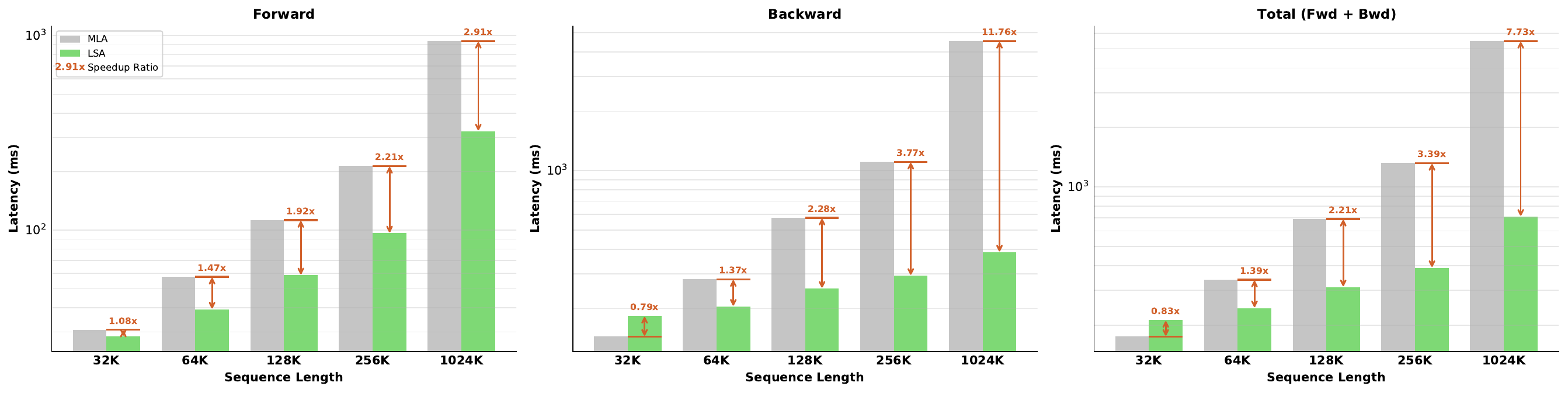}
    \caption{Single-attention-layer training latency of LSA and dense MLA across context lengths. Bars report forward, backward, and total latency, including kernel execution and CP communication; arrows indicate LSA speedup over MLA. The CP degrees are 4, 8, 16, 32, and 128 for context lengths of 32K, 64K, 128K, 256K, and 1024K, respectively.}
    \label{fig:train_layer_mla}
\end{figure}

\cref{sec:training_speedup} benchmarks LSA against the DSA sparse baseline. For completeness, \cref{fig:train_layer_mla} compares the per-layer training latency against dense MLA, highlighting a clear length-dependent crossover point. At short sequence lengths ($<64$K), dense MLA runs faster because LSA incurs indexing overheads (yielding a net latency of $0.83\times$ at 32K). However, beyond 64K, the quadratic scaling of dense MLA takes over—especially in the backward pass—allowing LSA to reverse the gap sharply, reaching up to a $7.73\times$ speedup at 1024K. This demonstrates that replacing dense MLA with LSA delivers rapidly compounding efficiency gains precisely for long-context regimes ($\ge64$K).

Note that these microbenchmarks are measured with fixed sequence lengths; when accounting for variable-length sequence packing in our actual data mixture, the practical efficiency crossover occurs at 128K.

\section{Serving setup and KV-cache partitioning}
\label[appendix]{app:serving_setup}

For long-context inference, the KV cache may exceed the memory capacity of a single accelerator. We therefore introduce KV-cache Partition (KVP), which shards cache pages across multiple ranks. Specifically, page $i$ is assigned to rank $i \bmod N_{\mathrm{KVP}}$, balancing memory usage across the KVP group. Our serving system uses length-aware routing to dispatch requests to instances with parallel configurations optimized for their context lengths.

\begin{itemize}
\item \textbf{Prefill.}
Prefill uses a 16K context chunk size with TP=8, EP=8, PP=2, and a CP/KVP group size of 8. The indexer K cache and attention KV cache are sharded across KVP ranks at page granularity. Before processing each chunk, the complete cache blocks are all-gathered so that every rank can attend to the full context.

\item \textbf{Decoding.}
Decoding uses a 4K chunk size. Short-context requests are served with data-parallel attention and expert parallelism (DP=16, EP=16). Requests with at least 256K context tokens are routed to two data-parallel replicas, each using an eight-rank KVP group.

With KVP, the indexer and SFA operate on local cache shards. Each rank selects its local Top-$K$ candidates; the resulting $K N_{\mathrm{KVP}}$ candidates are then all-gathered and re-ranked to obtain the global Top-$K$. In parallel, SFA computes attention over each local KV shard and combines the partial outputs using log-sum-exp normalization statistics, yielding the same result as attention over the global cache.
\end{itemize}

\clearpage
\section{Case Study: LSA Behavior on Needle-in-a-Haystack}
\label{app:casestudy}

To illustrate how LSA's sparse selection behaves, we present a detailed case study on the RULER NIAH (needle-in-a-haystack) multi-key task, which directly probes the indexer: retrieving a single \emph{needle} line from many near-identical distractors aligns precisely with the indexer's objective of isolating the few relevant tokens within a long, distractor-heavy context. Each prompt comprises a task description, a sequence of \texttt{key}--\texttt{value} lines (``One of the special magic uuids for \texttt{<key>} is: \texttt{<value>}''), and a closing question that queries the value of one specific key; the single needle line holds the answer (an abridged example is shown in \cref{box:cs_prompt}).

The model adopts two of the three LSA components. \emph{Streaming-Aware Indexing} partitions the budget into a fixed sink ($K_{\text{sink}}{=}16$), a sliding window ($K_{\text{swa}}{=}1024$), and a dynamic sparse set (a $\sim$1:1 fixed-to-sparse ratio); \emph{Cross-Layer Indexing} with group size $N{=}2$ lets each pair of consecutive layers share one indexer, so we label the shared figures by the owner layer (e.g.\ L26 for the L26--L27 pair). We omit \emph{Hierarchical Indexing}: it is a training-free, inference-time approximation that accelerates the selection rather than reshaping it during training, and thus lies outside the learned selection behavior studied here. For display clarity, all figures are computed on a single representative sample of length $S \approx 6$K.

\begin{figure}[ht]
\centering
\begin{tcolorbox}[colback=gray!3!white, colframe=black!60, arc=3pt, boxrule=0.8pt,
                  left=6pt, right=6pt, top=6pt, bottom=6pt, fontupper=\scriptsize\ttfamily,
                  width=\linewidth]
\colorbox{blue!14}{A special magic uuid is hidden within the following text. Make sure to memorize it. I will quiz you about the uuid}\\
\colorbox{blue!14}{afterwards.}\\
One of the special magic uuids for fb62b35d-...-03bd48468d49 is: fb2d6908-...-f1eaf31b7c54.\\
One of the special magic uuids for 1b20a57e-...-12df0ef04ec3 is: b364c1fd-...-6f649741c0fa.\\
\textcolor{gray}{\quad$\vdots$\quad (72 more distractor lines)}\\
\colorbox{red!14}{One of the special magic uuids for \textbf{e8ea1605-...-0cd28545ce6d} is: \textbf{9ac11a87-...-04225be0162e}.}\\
\textcolor{gray}{\quad$\vdots$\quad (3 more distractor lines)}\\
\colorbox{green!16}{What is the special magic uuid for e8ea1605-...-0cd28545ce6d mentioned in the provided text?}\\
\colorbox{green!16}{The special magic uuid for e8ea1605-...-0cd28545ce6d mentioned in the provided text is}\\[2pt]
\textbf{Target:} 9ac11a87-...-04225be0162e
\end{tcolorbox}
\caption{Abridged prompt for the case-study sample. The three spans analyzed in the visualizations below are color-coded: the \colorbox{blue!14}{task description}, the \colorbox{red!14}{target needle line} (which supplies the answer), and the \colorbox{green!16}{question}. The remaining 77 lines are same-format distractors; UUIDs are shortened for display.}
\label{box:cs_prompt}
\end{figure}

\subsection{Overview}
\label[appendix]{app:cs_overview}

\begin{figure}[ht]
    \centering
    \includegraphics[width=\linewidth]{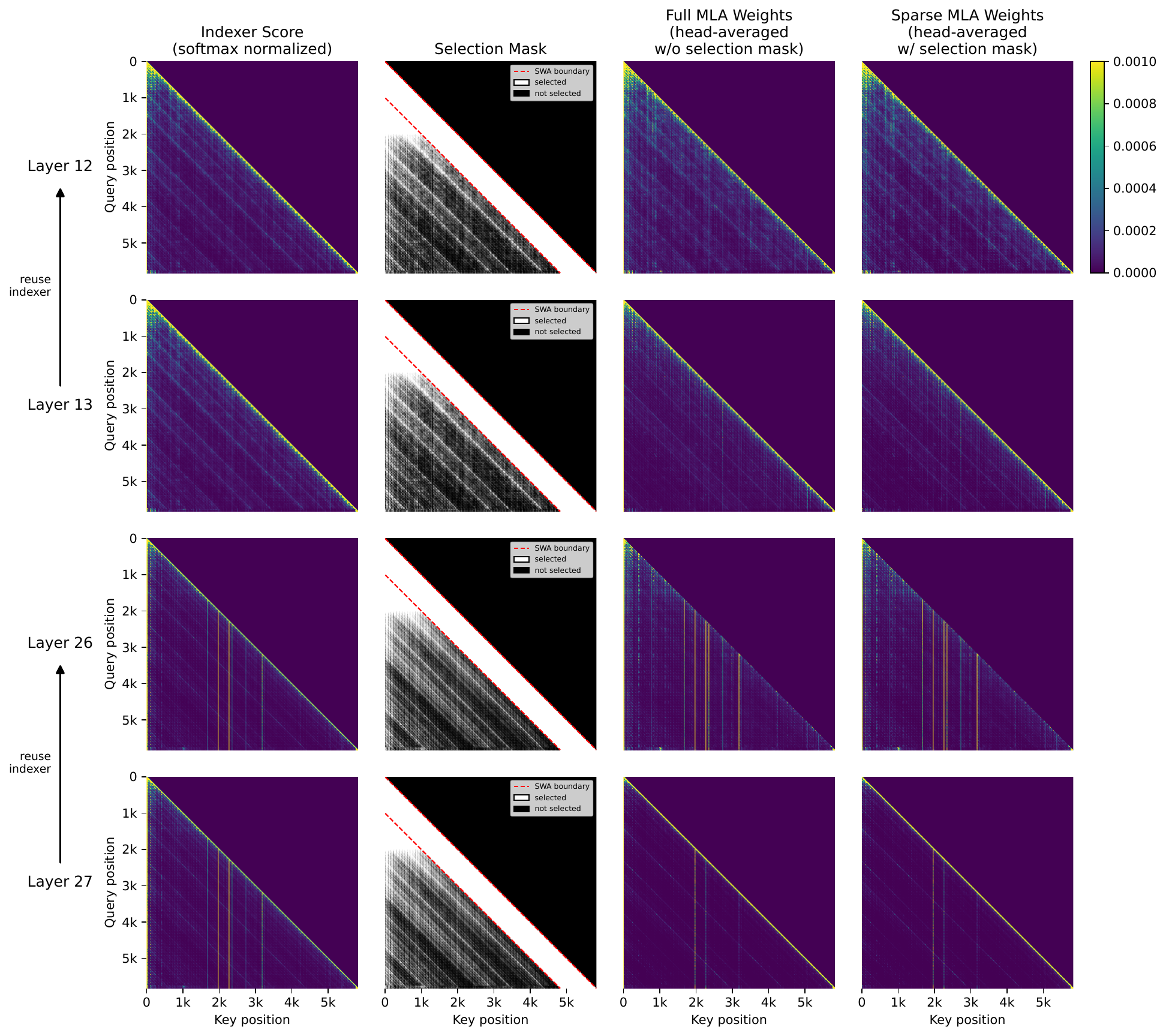}
    \caption{Overview of attention maps for the case-study sample at four layers (12, 13, 26, 27). Columns: (1) LSA indexer score (softmax-normalized), (2) selection mask (Top-$\kmax$, $\kmax{=}2048$; white = selected; red dashed lines mark sliding-window boundaries), (3) Full Attention (MLA, head-averaged), (4) LSA sparse attention (head-averaged, after selection).}
    \label{fig:cs_overview}
\end{figure}

\cref{fig:cs_overview} gives an overview spanning four quantities: the indexer score, the selection mask (the fixed sink and sliding-window positions of Streaming-Aware Indexing together with the indexer's Top-$\kmax$ choice), the full MLA attention weights ($\mathbf{p}$ in \cref{eq:warmup_loss}), and the resulting sparse MLA weights (the full weights after applying the selection mask). Several of these quantities are produced by a semantically equivalent reference implementation for analysis, rather than by the efficient fused-kernel implementation used in actual training and inference. The latter never instantiates a softmax over the indexer score, an explicit selection mask, or the full MLA weight matrix; we recompute all of them here purely for visualization, and normalize the indexer score by a softmax to make it comparable to the attention weights. For a CLI reuse layer, the indexer score and selection mask are copied directly from its group owner; for space we show two CLI groups, one from the middle and one from the tail of the network (layers 12/13 and 26/27). Full MLA weights, and sparse MLA weights are all clipped at a maximum of $10^{-3}$ for display.

Several patterns stand out. First, the indexer score closely tracks the full MLA weights, indicating that the KL distillation during training lets the indexer faithfully reproduce the core attention distribution. Second, the sparse MLA weights closely match the full MLA weights, showing that the post-selection attention is a good approximation of full attention. Third, the selection mask reserves a wide bright band along the diagonal, the fixed SWA region. Finally, the indexer score, full MLA weights, and sparse MLA weights all exhibit bright regions along the first column and the diagonal, indicating that attention mass naturally concentrates on the initial tokens and the local neighborhood---exactly the structure that motivates the fixed sink and sliding-window (SWA) budgets of Streaming-Aware Indexing. The following two subsections examine the discrete \emph{selection} (\cref{app:cs_selection}) and the continuous \emph{attention weights} (\cref{app:cs_weights}) in detail.

\clearpage
\subsection{Indexer Selection}
\label[appendix]{app:cs_selection}

\begin{figure}[ht]
    \centering
    \includegraphics[width=\linewidth]{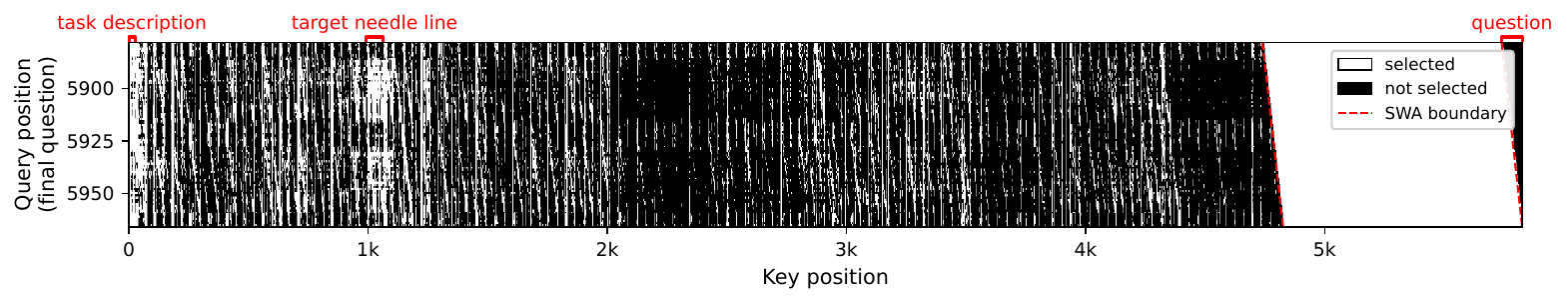}
    \caption{LSA indexer selection mask for final-question query tokens at layer 26, whose indices are shared with layer 27 via Cross-Layer Indexing (N=2). White and black denote selected and unselected positions, respectively. Red dashed lines indicate the causal and SWA boundaries, while red brackets highlight the task-description span, target needle line, and question key span.}
    \label{fig:cs_selection_query}
\end{figure}

\begin{figure}[ht]
    \centering
    \includegraphics[width=\linewidth]{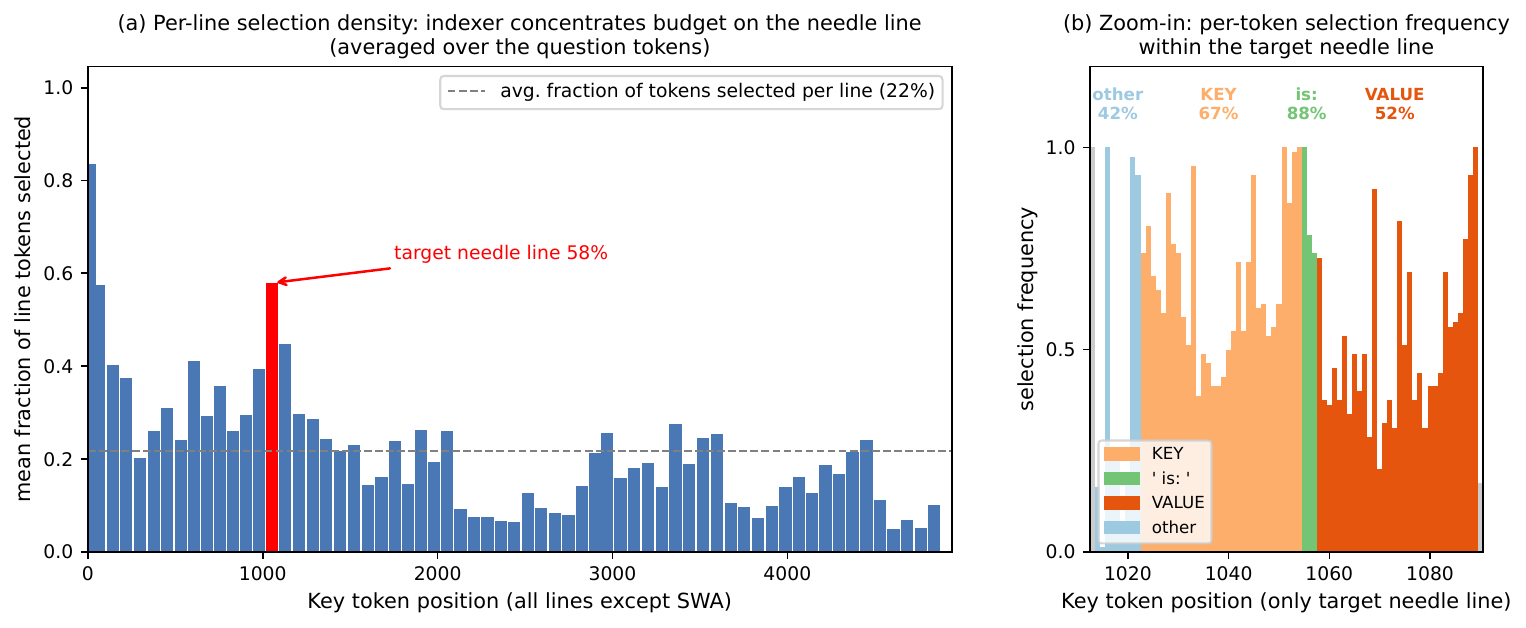}
    \caption{(a) Per-line selection density at layer 26: the fraction of each line's tokens selected by the indexer, averaged over the question tokens (lines fully covered by SWA are omitted). The target needle line is highlighted in red, and the dashed line denotes the average selection density across lines. (b) Per-token selection frequency within the target needle line, colored by segment (KEY / `` is: '' / VALUE / other). The needle line is selected far more often than distractor lines, and within it the KEY and VALUE tokens are preferentially selected---evidence that the indexer targets both the answer and its key.}
    \label{fig:cs_zoom}
\end{figure}

\begin{figure}[ht]
    \centering
    \includegraphics[width=\linewidth]{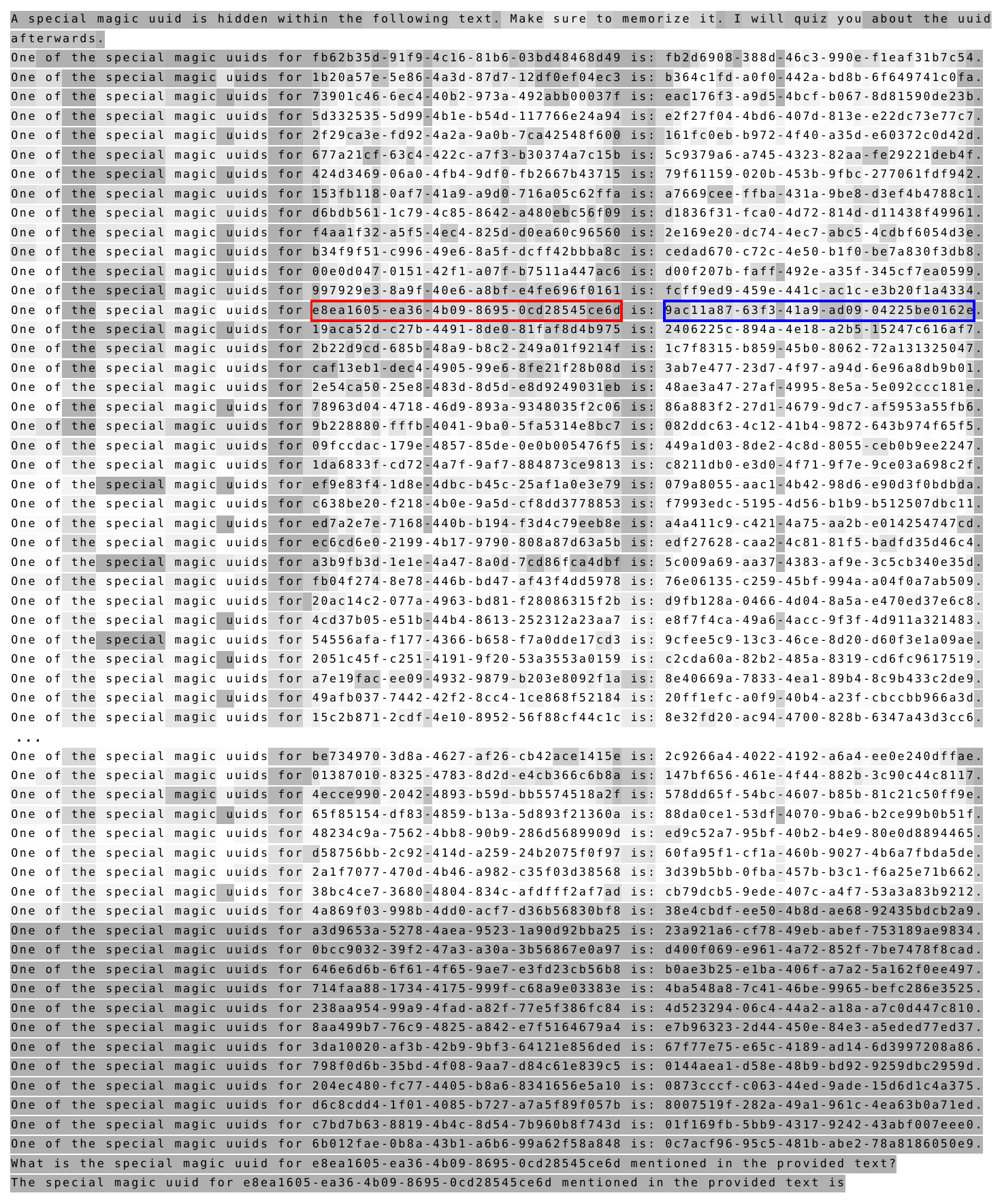}
    \caption{Per-token selection frequency (layer 26, shared with layer 27), averaged over the question queries, overlaid on the prompt text. Darker gray = more frequently selected. The middle distractor block is truncated (``...'') and the intro line is wrapped for layout; truncation does not affect the computation. Red/blue boxes mark the KEY/VALUE spans on the target needle line.}
    \label{fig:cs_text_select}
\end{figure}

We now focus on the discrete selection produced by the LSA indexer. \cref{fig:cs_selection_query} extracts the layer 26 selection mask from the \cref{fig:cs_overview}, using the question tokens as queries. Beyond the bright diagonal band at the tail corresponding to the fixed SWA region, two sharply localized high-selection regions emerge in the key sequence: one near the beginning of the context and another around position 1K. These regions correspond to the task description and the target needle line, respectively, showing that the indexer effectively identifies the information-bearing tokens required to answer the query while filtering out irrelevant distractors. \cref{fig:cs_zoom}(a) quantifies this behavior by measuring the fraction of tokens selected from each line by question queries (excluding positions covered by SWA). The target needle line is selected substantially more frequently than other lines, achieving a selection ratio of 58\%, compared with an average of 22\% across all lines. \cref{fig:cs_zoom}(b) further decomposes selection within the needle line: compared with the surrounding text, the KEY and VALUE tokens, as well as the connecting token ``is:'', receive significantly higher selection rates. \cref{fig:cs_text_select} visualizes the same pattern at token granularity by highlighting selected tokens directly in the prompt.

\FloatBarrier
\subsection{Attention Weights}
\begin{figure}[ht]
    \centering
    \includegraphics[width=\linewidth]{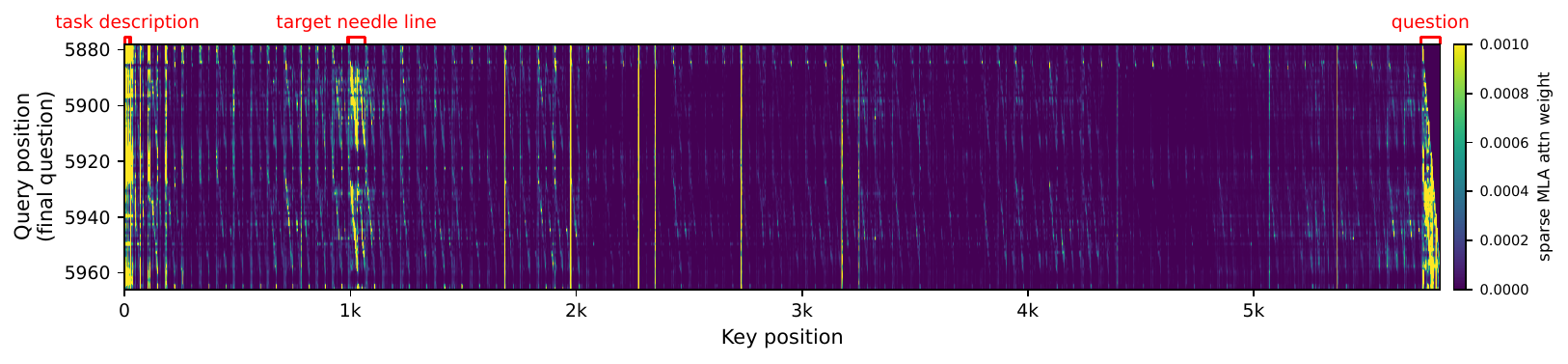}
    \caption{LSA sparse attention weights for the final-question query tokens (tail rows, 88 tokens) at layer 26. Color is the per-(query, key) attention weight, linearly clipped to $[0, 10^{-3}]$ to reveal weak long-range dependencies (the true maximum is ${\sim}0.26$). Red brackets above the panel mark the key spans of the task description, the target needle line, and the question. The question queries attend strongly to the task description, the target needle line, and the question span itself.}
    \label{fig:cs_sparse_query}
\end{figure}

\label[appendix]{app:cs_weights}
Mirroring the analysis above, we next examine the continuous attention weights after sparse selection. \cref{fig:cs_sparse_query} extracts the layer-26 sparse attention heatmap for question queries from the overview. In addition to the high-weight regions over the task description and target needle line already identified by the selection mask, the question span also emerges as a prominent attention region, which was previously hidden by the fixed SWA coverage.

\begin{figure}[ht]
    \centering
    \includegraphics[width=\linewidth]{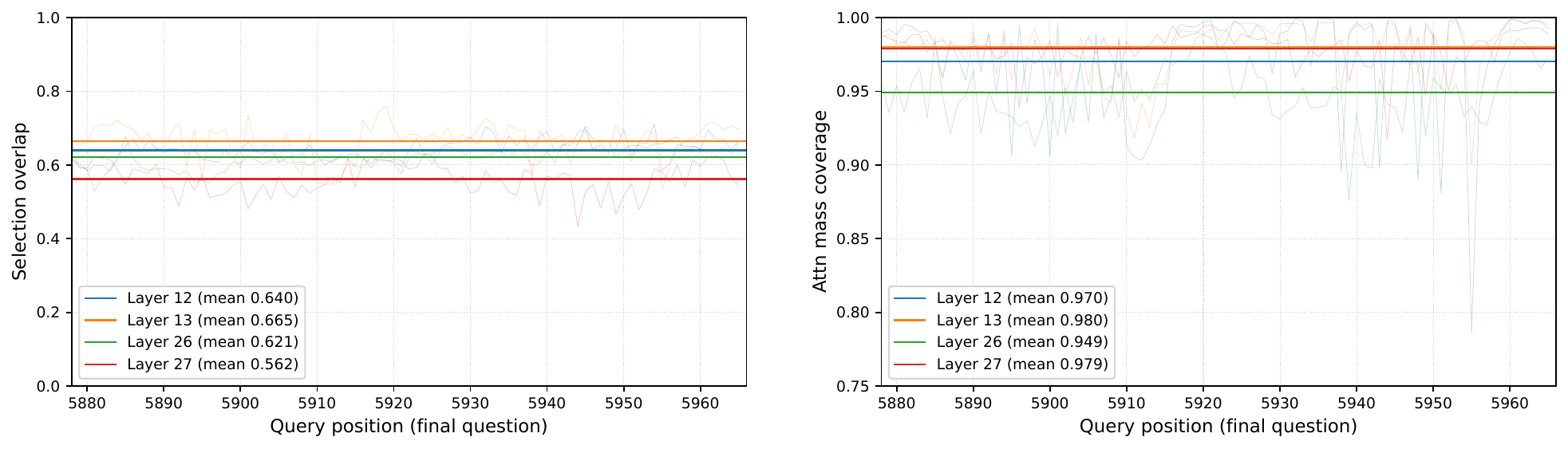}
    \caption{Sparse selection quality over the question tokens (tail queries) for layers 12/13/26/27. Left: Selection overlap between the indexer selection and the full-MLA Top-2048 within the causal range. Right: coverage, the fraction of dense attention mass retained by the selection. Thin curves are per-query values; horizontal lines are per-layer means. Here $\text{Overlap} = |\mathcal{S}_t \cap \mathcal{S}_t^{\text{MLA}}| / K$ with $K = 2048$ ($\mathcal{S}_t$ the indexer-selected set of \cref{eq:set_split} and $\mathcal{S}_t^{\text{MLA}}$ the Top-$K$ keys of the full-MLA row within the causal range), and $\text{Coverage} = \sum_{s \in \mathcal{S}_t} \alpha_{t,s}$ with $\alpha$ the head-averaged attention weight.}
    \label{fig:cs_coverage}
\end{figure}

\cref{fig:cs_coverage} quantifies how well sparse selection preserves dense attention. For each question query, we measure both the overlap between the selected tokens and the dense full-MLA Top-$\kmax$ set, and the coverage of the retained full-attention mass. Despite only moderate Top-$\kmax$ overlap ($\sim$0.56--0.66), the coverage remains consistently high ($\sim$0.95--0.98), indicating that sparse selection captures the few dominant keys that contribute most of the attention mass, even without exactly matching the dense Top-$\kmax$ set.

\begin{figure}[ht]
    \centering
    \includegraphics[width=\linewidth]{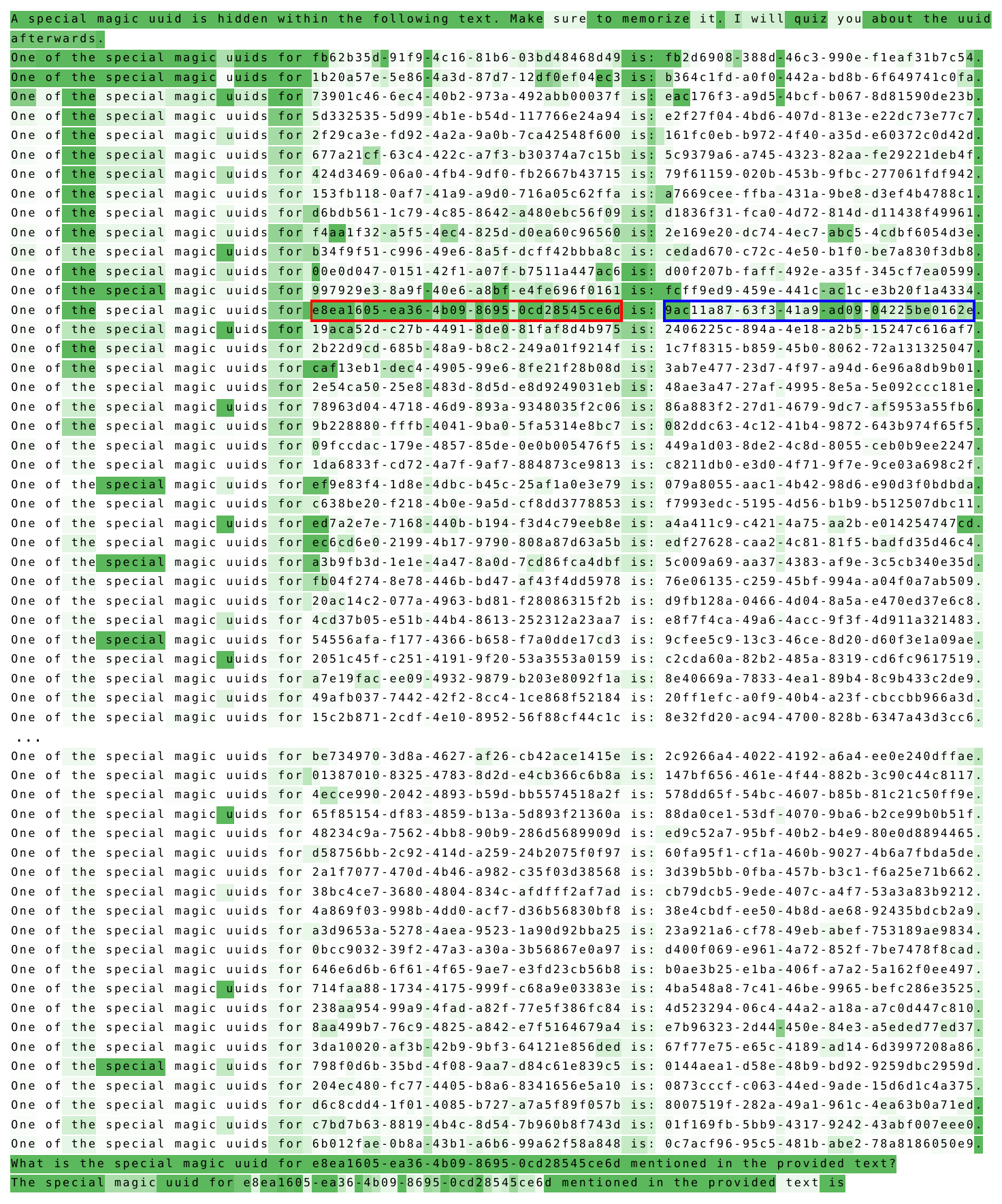}
    \caption{Per-token LSA sparse attention (layer 26), averaged over the question queries (causal-aware), overlaid on the prompt text. Greener = higher weight (clipped at $10^{-3}$). The middle distractor block is truncated (``...'') and the intro line is wrapped for layout; truncation does not affect the computation. Red/blue boxes mark the KEY/VALUE spans on the target needle line. Despite attending to only a small fraction of the context, LSA concentrates its attention mass on the needle's KEY and VALUE, confirming that the sparse indexer reliably locates the relevant tokens.}
    \label{fig:cs_text_sparse}
\end{figure}

This observation also supports CLI. Within each layer group (L12--L13 and L26--L27), the owner and reuse layers exhibit similar overlap (0.640 vs.\ 0.665; 0.621 vs.\ 0.562) and coverage (0.970 vs.\ 0.980; 0.949 vs.\ 0.979), demonstrating that a single shared index set can effectively serve both layers. Finally, \cref{fig:cs_text_sparse} visualizes the sparse attention distribution at token granularity by overlaying attention weights on the prompt text, revealing concentrated attention on the task description, question span, and, most prominently, the needle's KEY and VALUE tokens.

\end{document}